\documentclass[10pt,twocolumn,letterpaper]{article}
\usepackage[T1]{fontenc}
\usepackage[pagenumbers]{cvpr}
\usepackage{multirow}
\usepackage{array}
\usepackage{colortbl}
\usepackage{placeins}
\usepackage{algorithm}
\usepackage{algorithmic}
\usepackage{amsfonts,bm}
\usepackage{microtype}

\definecolor{promptframe}{RGB}{45,67,101}
\definecolor{promptback}{RGB}{244,247,250}
\definecolor{codeframe}{RGB}{80,80,80}
\definecolor{codeback}{RGB}{248,248,248}
\newsavebox{\pbbox}
\makeatletter
\newenvironment{promptbox}[1][]{%
  \def\pb@title{#1}\def\pb@frame{promptframe}\def\pb@back{promptback}%
  \par\medskip\noindent
  \begin{lrbox}{\pbbox}\begin{minipage}{\dimexpr\linewidth-12.8pt\relax}\vspace{2pt}%
}{%
  \vspace{1pt}\end{minipage}\end{lrbox}%
  \noindent{\setlength{\fboxsep}{4pt}\colorbox{\pb@frame}{\makebox[\dimexpr\linewidth-8pt\relax][l]{\color{white}\bfseries\footnotesize\pb@title}}}%
  \par\nopagebreak[4]\nointerlineskip\noindent
  {\setlength{\fboxsep}{6pt}\fcolorbox{\pb@frame}{\pb@back}{\usebox{\pbbox}}}%
  \par\medskip
}
\newenvironment{codebox}[1][]{%
  \def\pb@title{#1}%
  \par\medskip\noindent
  \begin{lrbox}{\pbbox}\begin{minipage}{\dimexpr\linewidth-12.8pt\relax}\vspace{2pt}%
}{%
  \vspace{1pt}\end{minipage}\end{lrbox}%
  \noindent{\setlength{\fboxsep}{4pt}\colorbox{codeframe}{\makebox[\dimexpr\linewidth-8pt\relax][l]{\color{white}\bfseries\footnotesize\pb@title}}}%
  \par\nopagebreak[4]\nointerlineskip\noindent
  {\setlength{\fboxsep}{6pt}\fcolorbox{codeframe}{codeback}{\usebox{\pbbox}}}%
  \par\medskip
}
\makeatother
\definecolor{headerblue}{RGB}{228,234,241}
\definecolor{oursgreen}{RGB}{231,240,238}
\newcommand{\cmark}{\textcolor{green!55!black}{\checkmark}}
\newcommand{\xmark}{\textcolor{red!65!black}{\ensuremath{\times}}}
\newcommand{\dataset}{CamChoreo}

\newcommand{\method}{CamDistill}
\newcommand{\inject}{CamInject}
\newcommand{\gcte}{\textsc{GCTE}}
\newcommand{\fcva}{frame-wise cross-attention}
\newcommand{\gcsa}{global camera self-attention}

\newcolumntype{L}[1]{>{\raggedright\arraybackslash}p{#1}}
\newcolumntype{C}[1]{>{\centering\arraybackslash}p{#1}}

\definecolor{cvprblue}{rgb}{0.21,0.49,0.74}
\usepackage[breaklinks,colorlinks,allcolors=cvprblue]{hyperref}
\hypersetup{
  pdftitle={CamChoreo: Understanding Compositional Camera Motion over Time},
  pdfauthor={Dazhao Du, Shiyan Du, Jian Liu, Yongjian Yu, Bohai Gu, Tao Han, Jie Zhang, Hualuo Liu, Eric Liu, Yujia Zhang, Xi Chen, Song Guo},
  pdfsubject={Public arXiv preprint},
  pdfkeywords={camera motion, video understanding, vision language models, benchmark, geometric distillation}
}
\title{CamChoreo: Understanding Compositional Camera Motion over Time}
\author{%
  Dazhao Du\textsuperscript{1}\thanks{Work done during an internship at Tencent.},\quad
  Shiyan Du\textsuperscript{2},\quad
  Jian Liu\textsuperscript{1},\quad
  Yongjian Yu\textsuperscript{2},\quad
  Bohai Gu\textsuperscript{1},\quad
  Tao Han\textsuperscript{1}\\[4pt]
  Jie Zhang\textsuperscript{1},\quad
  Hualuo Liu\textsuperscript{2},\quad
  Eric Liu\textsuperscript{2},\quad
  Yujia Zhang\textsuperscript{2},\quad
  Xi Chen\textsuperscript{2},\quad
  Song Guo\textsuperscript{1}\thanks{Corresponding author.}\\[9pt]
  {\normalfont\normalsize\textsuperscript{1}The Hong Kong University of Science and Technology\qquad
  \textsuperscript{2}Tencent}%
}
\makeatletter
\def\@maketitle{%
  \newpage\null\vspace*{8pt}%
  \begin{center}
    {\Large\bfseries\@title\par}
    \vspace{14pt}
    {\normalsize\bfseries\lineskip .5em
      \begin{tabular}[t]{c}\@author\end{tabular}\par}
    \vspace{9pt}
  \end{center}
}
\makeatother

\begin{document}
\maketitle
\begin{abstract}
Camera motion within a continuous shot can change over time and combine multiple movements, yet existing evaluations reduce it to clip-level labels. We study temporally grounded, compositional camera motion understanding. To our knowledge, this is the first task formulation requiring models to localize variable-length motion intervals and identify the complete set of direction-aware movements active in each interval. We introduce CamChoreo, a benchmark of 4{,}229 diverse single-shot clips with expert temporal annotations over 20 direction-aware labels, together with a separate training corpus of approximately 43K clips. Over half of the benchmark clips contain multiple intervals, and nearly half of the intervals contain simultaneous movements. Current vision-language models (VLMs) struggle on this task, while supervised fine-tuning (SFT) on our training corpus establishes strong baselines. To further improve recognition, we propose CamDistill, which transfers cross-frame geometric knowledge from a frozen 3D foundation model into a lightweight camera-token branch. The branch predicts per-frame camera representations from intermediate frozen visual features and aligns them with the teacher's camera tokens, removing the 3D model at inference. On CamChoreo, it improves frame-level micro F1 over same-data SFT by 5.4/3.7 points with 4B/8B backbones. It closely approaches CamInject, a direct geometric feature-injection baseline, while reducing inference latency by approximately 36\%. \method{}-8B also achieves strong zero-shot transfer to CameraBench and CameraMotionVQA. We will publicly release \dataset{}, the training corpus, and all models to support future research on temporally grounded, compositional camera motion understanding.
Project page: \url{https://ddz16.github.io/cammotion.github.io/}.
\end{abstract}

\begin{figure}[!t]
\centering
\includegraphics[width=\linewidth]{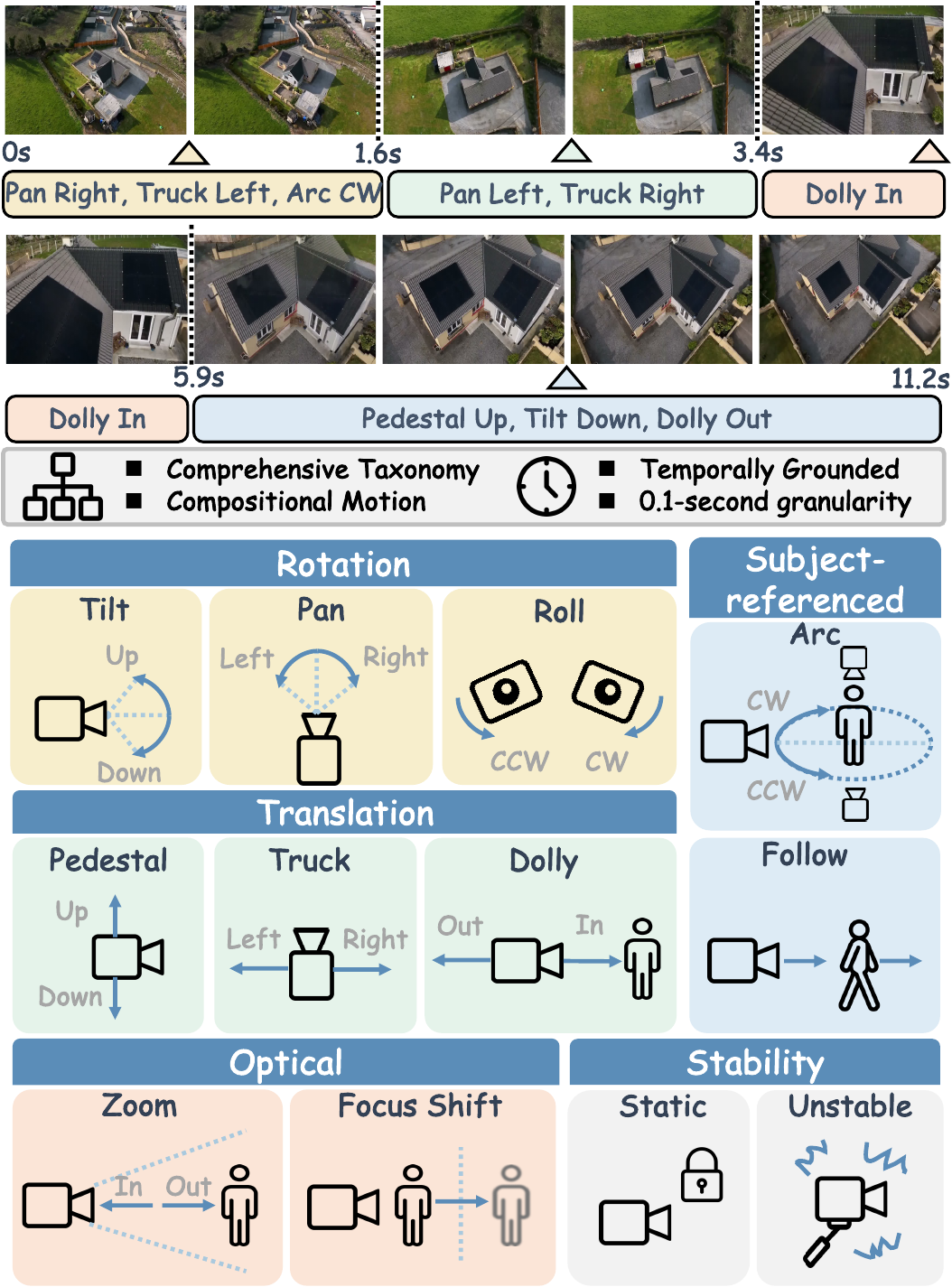}
\caption{\textbf{Temporally grounded compositional camera motion.} A shot is represented by motion-consistent intervals, each carrying all simultaneous movements. \dataset{} covers 12 types (20 direction-aware labels) grouped into five families: rotation, translation, optical, subject-referenced, and stability.}
\label{fig:task}
\end{figure}

\section{Introduction}
Camera motion shapes how a scene is observed and how visual information unfolds over time~\cite{spottiswoode1959grammar,yilmaz2023embodiment}. A pan redirects attention, a dolly changes viewpoint and reveals depth through parallax, and a zoom changes framing through optical magnification. Within a continuous shot, these movements may occur sequentially or simultaneously. Understanding camera behavior therefore requires recovering both their composition and temporal changes. Existing camera motion benchmarks typically assign a single label or a set of labels to an entire clip~\cite{lin2026towards,liu2026shotbench,feng2026geometry,tan2026natural}. Such labels indicate which movements are present, but cannot distinguish sequential from simultaneous movements or identify when the active motion changes. These limitations motivate treating camera motion as a time-varying, compositional signal rather than a fixed video attribute.

We introduce a new task: \emph{temporally grounded, compositional camera motion understanding}. Given a video, the model partitions each shot into motion-consistent intervals and predicts the complete set of direction-aware movements active in each interval. Figure~\ref{fig:task} illustrates the task: the top panel annotates one shot as a sequence of intervals, each carrying several simultaneous movements, and the bottom panel shows our taxonomy grouped into five families. The task thus asks \emph{what} the camera does, \emph{when} each movement occurs, and \emph{which} movements co-occur. To support it, we introduce \textbf{\dataset{}}, named for how a shot \emph{choreographs} camera-motion primitives over time. It contains 4{,}229 single-shot YouTube clips, covering 12 movement types and 20 direction-aware labels grounded in classical camera terminology \cite{nielsen2007camera}, with boundaries at 0.1-second resolution. Within this benchmark, temporal variation and compound motion are common: 2{,}411 clips contain multiple segments, and 3{,}797 segments contain compound camera motion, with multiple active labels under the annotation vocabulary. We also construct a separate training corpus of approximately 43K clips with the same taxonomy and interval-level supervision, enabling supervised learning.

Fine-tuning open-source VLMs such as Qwen3-VL \cite{qwen3vl} on our training corpus yields strong baselines for this task. To further improve these baselines, we complement the primarily semantic features of VLM visual encoders \cite{radford2021learning,tschannen2025siglip} with explicit geometric information, since understanding camera motion requires interpreting cross-frame cues such as parallax, perspective change, and horizon rotation. Specifically, we introduce \textbf{\inject{}}, which injects features from a frozen 3D foundation model \cite{wang2025vggt,wang2026vggt} into the VLM. \inject{} improves over SFT but requires running the costly 3D model at inference. We propose \textbf{\method{}} to retain the geometric benefit without this inference dependency. A lightweight Geometry-aware Camera Token Extractor (\gcte{}) predicts one camera token per frame from intermediate frozen visual features \cite{dosovitskiy2020image}. A distillation objective \cite{hinton2015distilling} aligns these tokens with the camera representation of a frozen 3D teacher, transferring geometric knowledge into the VLM during training. At inference, the 3D teacher is removed, leaving a compact geometry-aware stream with almost no runtime overhead.

Experiments show that SFT reaches 62.1/64.1 frame-level micro F1 on \dataset{} with Qwen3-VL 4B/8B backbones, outperforming advanced closed-source VLMs. \method{} further improves these results to 67.5/67.8, closely approaching \inject{} while reducing inference latency and memory usage. \method{}-8B also achieves strong zero-shot performance on the external camera motion understanding benchmarks CameraBench \cite{lin2026towards} and CameraMotionVQA \cite{feng2026geometry}.

Our contributions are threefold:
\begin{itemize}[leftmargin=1.4em]
    \item \textbf{A new task formulation.} We introduce temporally grounded, compositional camera motion understanding, to our knowledge the first task to jointly require within-shot motion interval localization and complete recognition of the direction-aware movements active in each interval.
    \item \textbf{Benchmark and training resources.} We introduce CamChoreo, with 4{,}229 expert-annotated clips, and a separate 43K-clip training corpus using the same annotation format. Frame-level and segment-level metrics jointly assess motion recognition, temporal localization, and complete movement-set recovery.
    \item \textbf{Efficient geometry distillation.} CamDistill transfers camera representations from a frozen 3D foundation model into a lightweight camera-token branch. It improves over same-data SFT and approaches direct geometric injection while reducing inference latency by approximately 36\%.
\end{itemize}

\begin{table*}[!t]
\centering
\caption{\textbf{Comparison of camera motion and cinematography benchmarks.} ``Camera-specific'' indicates a benchmark dedicated to camera motion understanding. ``Temporal'' indicates explicit evaluation of motion interval boundaries and complete per-interval label sets. ``\# Classes'' counts direction-aware labels for \dataset{}.}
\label{tab:benchmark_compare}
\small
\setlength{\tabcolsep}{8.0pt}
\renewcommand{\arraystretch}{1.02}
\begin{tabular}{l l r c c c c c}
\toprule
Benchmark & Source & \# Clips & \# Classes & Real & Multi-label & Camera-specific & Temporal \\
\midrule
Cinematic2K \cite{Cinematic2K} & Web & 2{,}000 & 11 & \cmark & \xmark & \xmark & \xmark \\
VidComposition \cite{Vidcomposition} & Movies & 982 & 7 & \cmark & \cmark & \xmark & \xmark \\
CineTechBench \cite{wang2026cinetechbench} & Movies & 120 & 15 & \cmark & \cmark & \xmark & \xmark \\
ShotBench \cite{liu2026shotbench} & Movies & 464 & 16 & \cmark & \xmark & \xmark & \xmark \\
CameraBench \cite{lin2026towards} & Web & 3{,}381 & 23 & \cmark & \cmark & \cmark & \xmark \\
CameraMotionVQA \cite{feng2026geometry} & Synthetic & 12{,}274 & 15 & \xmark & \cmark & \cmark & \xmark \\
\midrule
\rowcolor{oursgreen}
\dataset{} (ours) & Web & 4{,}229 & 20 & \cmark & \cmark & \cmark & \cmark \\
\bottomrule
\end{tabular}
\end{table*}

\section{Related Work}
\subsection{Camera Motion Benchmarks}
Film theory has long studied camera movement as a device with narrative and emotional functions \cite{spottiswoode1959grammar,nielsen2007camera,yilmaz2023embodiment}. Recent cinematography benchmarks evaluate camera movement alongside shot scale, lighting, and composition, mainly through clip-level classification or multiple-choice QA \cite{Cinematic2K,Vidcomposition,wang2026cinetechbench,liu2026shotbench,wu2025refineshot}. CameraBench \cite{lin2026towards} introduces a broad expert vocabulary for real videos, while CameraMotionVQA \cite{feng2026geometry} supports controlled multi-label recognition on one-second synthetic clips. ACaM \cite{tan2026natural} evaluates atomic camera movements in real and synthetic videos. These benchmarks neither localize variable-length intervals within real shots nor evaluate how movements co-occur over time. \dataset{} is designed specifically for this temporally grounded, compositional setting, as summarized in Table~\ref{tab:benchmark_compare}.

\subsection{VLMs for Camera Motion Understanding}
General-purpose video VLMs \cite{llavaonevision,llavavideo,qwen3vl,internvl3} provide strong semantic understanding but are not trained to perceive camera geometry. Our task also relates to video temporal grounding, in which VLMs localize events along a timeline \cite{wu2025survey}. Recent camera-motion methods introduce structured reasoning traces, explicit pose grounding, or textual pose prompts derived from geometry models \cite{wu2026camreasoner,yang2026cambrian,feng2026geometry}. Other work augments VLMs with 3D priors \cite{zheng2026learning}. These studies demonstrate the value of geometry, but either retain the geometry model at inference or compress its output into discrete text. \method{} instead transfers the teacher's \emph{continuous} camera representation during training and removes the geometry model at inference.

\subsection{Camera Pose Estimation in Video}
Classical SfM and SLAM recover camera trajectories through feature matching and geometric optimization \cite{schonberger2016structure,davison2007monoslam,engel2014lsd,teed2021droid,li2026droid,zhang2022structure}. More recent feed-forward models such as DUSt3R, VGGT, and VGGT-$\Omega$ jointly predict camera pose and 3D scene structure in a single pass \cite{wang2024dust3r,wang2025vggt,wang2026vggt}, with further estimators improving robustness and multi-view consistency \cite{huang2025vipe,wang2026pi}. VGGT and VGGT-$\Omega$ in particular attach a dedicated camera token to each frame, from which that frame's camera pose can be decoded. We distill the camera token itself into the VLM, transferring a pose-associated, geometry-informed representation while leaving the mapping to camera-motion labels to the language model.

\section{\dataset{} Dataset}
\label{sec:benchmark}
\subsection{Task Definition}
Given a single-shot video (a continuous shot without transitions), the model divides it into a sequence of intervals with consistent camera motion. For each interval, it predicts the start and end timestamps and the complete set of camera movements active during that interval. The vocabulary contains 20 direction-aware labels derived from 12 movement types (Figure~\ref{fig:task}; Appendix~\ref{app:taxonomy}). A correct prediction must therefore recover both the segment boundaries and the complete set of co-occurring labels within each segment.

\begin{figure*}[!t]
\centering
\includegraphics[width=0.95\linewidth,trim=0 7bp 0 6.5bp,clip]{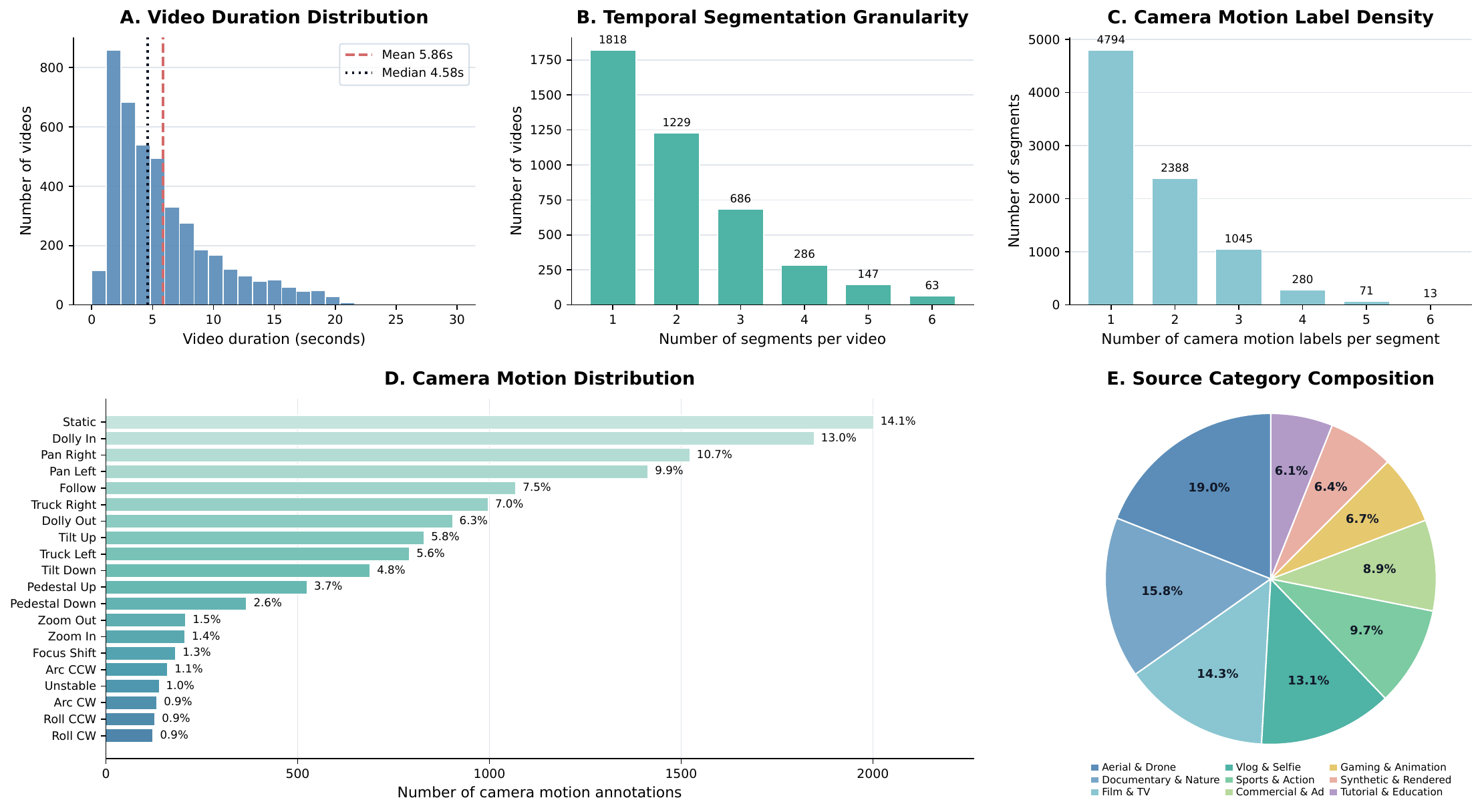}
\caption{\textbf{\dataset{} statistics.} \textbf{(A)} video duration, \textbf{(B)} number of segments per clip, \textbf{(C)} number of simultaneous movements per segment, \textbf{(D)} the long-tailed distribution of the 20 direction-aware labels, and \textbf{(E)} the nine content domains.}
\label{fig:data_stats}
\end{figure*}

\subsection{Benchmark Curation}
\label{sec:curation}
\noindent\textbf{Collection and filtering.}
We collect YouTube footage from nine content domains and split each video into single shots at hard cuts with TransNetV2 \cite{soucek2024transnet}. The resulting shots are filtered by duration and visual quality and de-duplicated, yielding a diverse clip pool. To surface rare motions, a preliminary model assigns pseudo labels that are used only to guide candidate sampling and never enter the final annotations. The full funnel is detailed in Appendix~\ref{app:pipeline}.

\noindent\textbf{Expert annotation.}
A team of five annotators with film- and media-related backgrounds label every clip from scratch, marking motion-consistent intervals and all active direction-aware movements. They use parallax to separate rotation from translation, perspective change to distinguish Dolly from Zoom, and scene context to separate camera from subject motion. Boundaries are placed at 0.1-second resolution wherever the active motion set changes. Annotations are cross-checked for quality. Clips with unresolved disagreements are discarded, leaving 4{,}229 videos in the final benchmark. Details are given in Appendix~\ref{app:annotation}.

\subsection{Training Corpus}
\label{sec:training_corpus}
To support supervised learning, we crawl videos covering diverse content from Tencent Video and construct a separate training corpus. We follow the same processing and expert annotation pipeline as the benchmark (Section~\ref{sec:curation}), including shot segmentation, filtering, duplicate removal, temporal annotation, and cross-review. The resulting corpus contains approximately 43K carefully annotated clips with the same camera motion taxonomy and annotation format.

\subsection{Benchmark Statistics}
\dataset{} contains 4{,}229 single-shot clips totaling 6.88 hours, with 8{,}591 expert-annotated segments and 14{,}258 movement instances. The average clip duration is 5.9 seconds, and the duration distribution is shown in Figure~\ref{fig:data_stats}A. Figure~\ref{fig:data_stats} summarizes the benchmark statistics.

\noindent\textbf{Temporal structure.}
Camera motion changes within most clips. As shown in Figure~\ref{fig:data_stats}B, 2{,}411 of the 4{,}229 clips contain multiple segments, with some containing as many as six. Clip-level annotation would therefore merge distinct motion phases in more than half of the benchmark.

\noindent\textbf{Compound camera motion.}
In the curated benchmark, 3{,}797 of the 8{,}591 segments (44.2\%) contain compound camera motion with at least two active movement labels, and some contain three or more (Figure~\ref{fig:data_stats}C). A single-label prediction therefore drops part of the active camera state in nearly half of the benchmark segments.

\noindent\textbf{Long-tailed labels and domains.}
The 20 direction-aware labels follow a pronounced long-tailed distribution (Figure~\ref{fig:data_stats}D). Static, Dolly, and Pan are frequent, whereas Zoom, Roll, and Focus Shift are rare. The clips span nine content domains, with aerial, documentary, and film footage contributing the largest shares (Figure~\ref{fig:data_stats}E). 

\noindent\textbf{Comparison with other benchmarks.}
Table~\ref{tab:benchmark_compare} highlights two limitations of existing benchmarks. General cinematography datasets cover camera movement only as one attribute among many, while camera-specific benchmarks provide richer motion vocabularies but still assign a single label set to an entire clip. Consequently, the representative benchmarks in Table~\ref{tab:benchmark_compare} do not capture how the active camera motion changes within a real shot. \dataset{} addresses this gap by combining real web video with camera-specific, direction-aware multi-label annotations over variable-length temporal segments. It therefore evaluates both compound camera motion within each segment and its evolution over time.

\begin{figure*}[!t]
\centering
\includegraphics[width=0.93\linewidth]{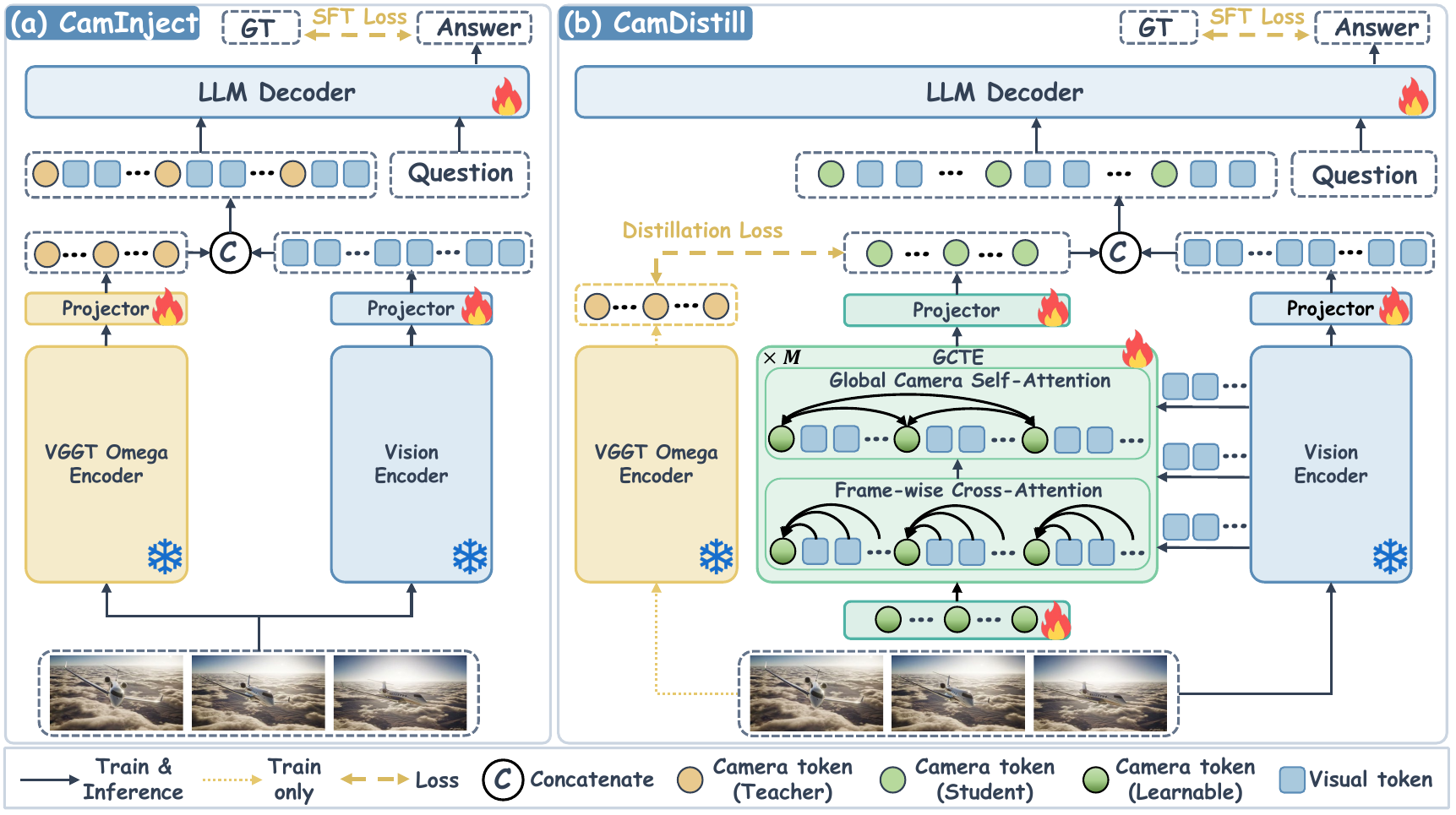}
\caption{\textbf{\inject{} versus \method{}.} \textbf{(a)} \inject{} runs a frozen 3D foundation model (VGGT-$\Omega$) beside the vision encoder and inserts each frame's \emph{teacher} camera token just before that frame's visual tokens in the LLM input, so the 3D model is required at inference. \textbf{(b)} \method{} instead trains a lightweight \gcte{}, a stack of alternating frame-wise cross-attention and global camera self-attention blocks, to predict the \emph{student} camera tokens, which are inserted at the same positions. A distillation loss aligns them with the teacher, which is removed at inference. Snowflake and flame icons denote frozen and trainable components.}
\label{fig:method}
\end{figure*}

\section{CamInject and CamDistill}
\label{sec:method}

\subsection{Motivation}
Recognizing camera motion requires comparing perspective, parallax, scale, and orientation across frames. VLM vision encoders, however, are optimized for semantic alignment and encode these geometric signals only weakly. Feed-forward 3D foundation models such as VGGT \cite{wang2025vggt} and VGGT-$\Omega$ \cite{wang2026vggt} are designed to recover them. Given a set of frames, these models estimate depth, point maps, and camera pose. In particular, they associate each frame with a dedicated \emph{camera token} from which its pose is decoded. The evolution of these tokens across time therefore provides a compact, camera-oriented geometric representation.

A direct way to exploit this signal is \inject{} (Figure~\ref{fig:method}a), which runs the 3D model alongside the frozen vision encoder, projects each teacher camera token into the LLM hidden space, and concatenates the projected tokens with the visual sequence. This baseline is effective (Section~\ref{sec:experiments}) but expensive: the 3D model must process every video at inference, and models such as VGGT-$\Omega$ apply global attention over the tokens of all frames, so latency and memory grow rapidly with video length. Since these camera tokens are clearly useful, we ask whether they can be obtained without running the 3D model at test time. \method{} moves the teacher entirely to training (Figure~\ref{fig:method}b). A lightweight student predicts per-frame camera tokens from the frozen vision features the VLM already computes, and a distillation loss aligns them with the teacher's tokens. The 3D model is then discarded, so \method{} keeps the geometric supervision with almost no inference overhead.

\subsection{Architecture}
The student takes the form of a \emph{Geometry-aware Camera Token Extractor} (\gcte{}), a lightweight branch attached to the frozen vision encoder (Figure~\ref{fig:method}b). It reads intermediate visual features without modifying the pretrained visual stream, produces one camera token per frame through alternating attention blocks, and places these tokens before the corresponding visual tokens. The LLM can then condition its predictions on both semantic visual features and an explicit camera representation.

\paragraph{Camera tokens.}
Let $x_i^{(\ell)}$ denote the frozen visual features of frame $i$ at encoder layer $\ell$, with $P_i$ visual tokens. \gcte{} reads \emph{intermediate} rather than final-layer features. Intermediate layers retain local geometric cues such as parallax and perspective change, while higher layers become increasingly specialized for semantic alignment \cite{feng2026geometry}. This choice gives the student access to a cleaner camera-related signal.

For a video with $T$ frames, \gcte{} maintains one camera token $c_i$ per frame, initialized with a learnable embedding $c_i^{(0)}$. Following the reference-view convention of 3D reconstruction, the first frame uses a dedicated embedding and all other frames share a second one, marking the first frame as the reference view. The attention blocks below then refine these tokens.

\paragraph{Alternating attention blocks.}
\gcte{} stacks $M$ alternating blocks, each a \emph{frame-wise cross-attention} followed by a \emph{global camera self-attention} (Figure~\ref{fig:method}b), mapping the initial tokens $\{c_i^{(0)}\}_{i=1}^{T}$ to the final camera states $\{c_i^{(M)}\}_{i=1}^{T}$. We write $c_i^{(m)}$ for the camera token of frame $i$ after block $m$. The two attention operations serve complementary roles. Frame-wise cross-attention extracts camera-relevant evidence from each frame, while global self-attention compares that evidence across time. Together, they represent both the geometry of individual views and its temporal evolution.

\paragraph{Frame-wise cross-attention.}
Frame-wise cross-attention lets each camera token read only the visual tokens from its corresponding frame:
\begin{equation}
\tilde{c}_i^{(m)} = \mathrm{CrossAttn}\big(c_i^{(m-1)},\ x_i^{(\ell_m)}\big).
\end{equation}
Here the camera token $c_i^{(m-1)}$ from the previous block is the query, the frame features $x_i^{(\ell_m)}$ at the layer $\ell_m$ tapped by block $m$ are the keys and values, and $\tilde{c}_i^{(m)}$ is the resulting camera token. This interaction is one-way: the block updates only the camera token. The visual tokens remain unchanged, preserving the feature distribution expected by the pretrained VLM.

\paragraph{Global camera self-attention.}
Global camera self-attention then allows the per-frame camera tokens to exchange information. Each token can therefore interpret its frame relative to the surrounding viewpoints rather than in isolation:
\begin{equation}
\bigl[c_i^{(m)}\bigr]_{i=1}^{T} = \mathrm{SelfAttn}\!\left(\bigl[\tilde{c}_i^{(m)}\bigr]_{i=1}^{T}\right).
\end{equation}
This operation attends over only the $T$ camera tokens, not the full set of $\sum_i P_i$ visual patches. Both attention modules use standard pre-norm transformer blocks \cite{vaswani2017attention,ba2016layer}.

\paragraph{Output and injection.}
For frame $i$, we form the student representation $z_i=[\tilde{c}_i^{(M)};c_i^{(M)}]$ by concatenating the final frame-level and temporally contextualized states. A two-layer MLP projects this representation to the LLM hidden size. The projected token is placed immediately before the visual tokens of frame $i$, making the camera representation available as context before the decoder processes the frame content. Appendix~\ref{app:gcte} provides the complete block specification.

\subsection{Distillation Objective}
\method{} is trained with two objectives (Figure~\ref{fig:method}b): the standard next-token prediction loss $\mathcal{L}_{\mathrm{SFT}}$ for the structured task output, and a camera-token \emph{distillation loss} $\mathcal{L}_{\mathrm{cam}}$ \cite{hinton2015distilling} that aligns each student token $z_i$ with the teacher's target camera token $g_i$ for frame $i$ by cosine distance:
\begin{equation}
\begin{aligned}
\mathcal{L}
&= \mathcal{L}_{\mathrm{SFT}} + \lambda_{\mathrm{cam}}\,\mathcal{L}_{\mathrm{cam}} \\
&= \mathcal{L}_{\mathrm{SFT}} + \frac{\lambda_{\mathrm{cam}}}{T}
   \sum_{i=1}^{T}\big(1-\cos(z_i,g_i)\big),
\end{aligned}
\end{equation}
where $\lambda_{\mathrm{cam}}$ weights the distillation term. Through $\mathcal{L}_{\mathrm{cam}}$, the student tokens are encouraged to reproduce the teacher's pose-associated camera representation.

\section{Experiments}
\label{sec:experiments}
\subsection{Setup}
\paragraph{Models.}
We evaluate all models on \dataset{} using the same instruction and output format. The comparison includes four groups. \emph{Closed-source APIs} comprise GPT-5.4 \cite{gpt5} and Gemini-3.1-Pro \cite{gemini3}. \emph{Open-source VLMs} include Qwen2.5-VL \cite{qwen25vl}, Qwen3-VL at 4B, 8B, and 235B \cite{qwen3vl}, Qwen3.5 \cite{qwen3.5}, Qwen3.6 \cite{qwen3.6-35b-a3b}, InternVL3.5 \cite{internvl3}, and Cam-Motion-7B, a Qwen2.5-VL model fine-tuned on CameraBench \cite{lin2026towards}. The \emph{geometry-only baseline} estimates per-frame camera pose with VGGT-$\Omega$ and maps translational and angular velocities to labels using hand-designed rules (Appendix~\ref{app:geobaseline}). Finally, \emph{our models} are \method{} and the direct-injection reference \inject{}, implemented with Qwen3-VL 4B and 8B backbones and VGGT-$\Omega$ as the default 3D teacher. For both methods, we fully fine-tune the language model while freezing the vision encoder, and jointly train \gcte{} for \method{}. The SFT baseline likewise fully fine-tunes the language model while freezing the vision encoder, but uses neither the \gcte{} module nor the distillation loss. All three methods are trained on our corpus of approximately 43K clips (Section~\ref{sec:training_corpus}). Appendix~\ref{app:implementation} provides complete training details.

\paragraph{Metrics.}
We evaluate predictions from two complementary perspectives (Appendix~\ref{app:metrics}). \emph{Frame-level} evaluation samples the timeline every 0.1\,s and computes precision, recall, and F1 over the 20 direction-aware labels. We report both micro averages, which weight instances equally, and macro averages, which weight classes equally. These metrics evaluate classification performance at each timestamp. \emph{Segment-level} evaluation instead matches predicted and ground-truth intervals by temporal IoU and reports F1 at thresholds $0.3/0.5/0.7$. Segment localization (SegLoc) evaluates temporal overlap without considering labels, whereas segment detection (SegDet) additionally requires an exact match of the direction-aware label set.

\begin{table*}[t]
\centering
\caption{\textbf{Results on \dataset{}.} Frame-level micro/macro precision (P), recall (R), and F1 are evaluated every 0.1 seconds. SegLoc measures temporal overlap; SegDet additionally requires an exact direction-aware label set. Both are reported at IoU $0.3/0.5/0.7$. Per column, best is in \textbf{bold} and second-best is \underline{underlined}.}
\label{tab:main_results}
\small
\setlength{\tabcolsep}{8.69pt}
\renewcommand{\arraystretch}{1.02}
\begin{tabular}{l *{12}{r}}
\toprule
 & \multicolumn{6}{c}{Frame-Level (\%)} & \multicolumn{6}{c}{Segment-Level F1 (\%)} \\
 & \multicolumn{3}{c}{Micro} & \multicolumn{3}{c}{Macro} & \multicolumn{3}{c}{SegLoc @IoU} & \multicolumn{3}{c}{SegDet @IoU} \\
\cmidrule(lr){2-4}\cmidrule(lr){5-7}\cmidrule(lr){8-10}\cmidrule(lr){11-13}
Model & P & R & F1 & P & R & F1 & 0.3 & 0.5 & 0.7 & 0.3 & 0.5 & 0.7 \\
\midrule
VGGT-$\Omega$ & 21.7 & 30.2 & 25.3 & 23.3 & 26.2 & 17.2 & 73.6 & 63.9 & 48.5 & 3.0 & 2.9 & 2.7 \\
\midrule
Gemini-3.1-Pro & 51.4 & 37.4 & 43.3 & 40.5 & 23.0 & 27.4 & 82.8 & 76.1 & 61.2 & 25.1 & 23.7 & 19.8 \\
GPT-5.4 & 43.8 & 33.5 & 38.0 & 31.2 & 19.2 & 21.3 & 82.8 & 76.1 & 62.1 & 23.9 & 22.4 & 18.9 \\
\midrule
InternVL3.5-8B & 31.9 & 18.7 & 23.6 & 16.9 & 6.6 & 7.1 & 38.4 & 21.0 & 8.5 & 7.2 & 4.0 & 1.7 \\
Qwen2.5-VL-7B & 27.9 & 17.8 & 21.8 & 8.4 & 4.1 & 4.0 & 62.6 & 39.7 & 18.2 & 9.9 & 6.4 & 3.0 \\
Cam-Motion-7B & 12.6 & 11.2 & 11.9 & 3.4 & 2.4 & 2.6 & 53.2 & 38.0 & 20.3 & 0.0 & 0.0 & 0.0 \\
Qwen3-VL-4B & 32.4 & 19.3 & 24.2 & 14.9 & 7.1 & 7.8 & 69.7 & 55.6 & 38.8 & 14.4 & 11.6 & 8.6 \\
Qwen3-VL-8B & 36.6 & 23.0 & 28.3 & 17.3 & 7.9 & 9.1 & 76.2 & 67.7 & 52.5 & 16.5 & 15.0 & 12.5 \\
Qwen3-VL-235B & 39.7 & 28.8 & 33.4 & 22.0 & 13.9 & 15.4 & 79.1 & 66.1 & 47.8 & 17.9 & 15.6 & 12.0 \\
Qwen3.5-4B & 46.0 & 14.7 & 22.3 & 20.6 & 5.1 & 6.9 & 52.1 & 46.9 & 38.2 & 16.8 & 15.4 & 12.9 \\
Qwen3.5-9B & 50.5 & 13.7 & 21.5 & 21.9 & 4.5 & 6.5 & 43.8 & 38.3 & 29.7 & 14.9 & 13.3 & 10.7 \\
Qwen3.6-35B & 46.4 & 29.3 & 35.9 & 22.3 & 12.2 & 14.3 & 80.1 & 68.8 & 51.2 & 19.8 & 17.9 & 14.2 \\
\midrule
SFT-4B & 67.7 & 57.3 & 62.1 & 53.1 & 46.9 & 49.0 & 85.2 & 78.9 & 64.7 & 35.1 & 33.6 & 29.2 \\
\rowcolor{oursgreen}
\method{}-4B & 73.5 & 62.4 & 67.5 & \underline{63.8} & 54.5 & 57.7 & 86.4 & \underline{80.4} & \underline{66.8} & 39.7 & 38.2 & 33.5 \\
\rowcolor{oursgreen}
\inject{}-4B & 73.0 & \underline{62.8} & 67.5 & 63.5 & 54.7 & 57.6 & \textbf{86.7} & \textbf{80.8} & \underline{66.8} & 40.1 & 38.6 & 33.9 \\
\midrule
SFT-8B & 69.5 & 59.4 & 64.1 & 59.6 & 49.5 & 51.9 & 85.9 & 79.7 & 65.3 & 36.5 & 34.9 & 30.5 \\
\rowcolor{oursgreen}
\method{}-8B & \underline{73.8} & 62.6 & \underline{67.8} & 63.6 & \underline{54.9} & \underline{57.9} & \underline{86.5} & \underline{80.4} & 66.7 & \textbf{40.5} & \underline{38.8} & \underline{34.1} \\
\rowcolor{oursgreen}
\inject{}-8B & \textbf{74.0} & \textbf{63.5} & \textbf{68.3} & \textbf{64.9} & \textbf{56.3} & \textbf{59.2} & \underline{86.5} & \textbf{80.8} & \textbf{67.0} & \underline{40.3} & \textbf{39.0} & \textbf{34.2} \\
\bottomrule
\end{tabular}
\end{table*}

\subsection{Main Results}
Table~\ref{tab:main_results} first reveals the limitations of relying on geometry alone. The VGGT-$\Omega$ pose-based baseline reaches 25.3 micro F1 and 63.9 SegLoc at IoU~0.5, but only 2.9 SegDet. Pose estimates help identify changes in camera behavior, while the hand-designed mapping from poses to labels struggles to recover complete movement sets and cannot recognize optical categories such as Zoom and Focus Shift. Open-source VLMs also struggle with this task. Cam-Motion-7B, despite its camera-motion-specific training, obtains 11.9 micro F1 and 38.0 SegLoc@0.5 on \dataset{}. General-purpose Qwen2.5-VL-7B and InternVL3.5-8B reach 21.8 and 23.6 micro F1, respectively. Temporal localization is also weak for some open-source models: their SegLoc@0.5 scores are only 39.7 and 21.0, respectively, indicating difficulty recovering motion interval boundaries. Scaling Qwen3-VL from 4B through 8B to 235B raises micro F1 from 24.2 through 28.3 to 33.4, showing that substantially greater model capacity brings only a modest improvement. Closed-source VLMs generally localize motion intervals more accurately, with both GPT-5.4 and Gemini-3.1-Pro reaching 76.1 SegLoc@0.5. They also perform better in recognition: GPT-5.4 reaches 38.0 micro F1, while Gemini-3.1-Pro, the strongest evaluated closed-source model, reaches 43.3. Even Gemini obtains only 23.7 SegDet@0.5, compared with 76.1 SegLoc@0.5, highlighting the difficulty of recognizing every active motion within a localized interval.

Fine-tuning Qwen3-VL-4B and Qwen3-VL-8B on our corpus of approximately 43K clips yields strong SFT baselines, reaching 62.1 and 64.1 micro F1. These scores exceed Gemini-3.1-Pro by 18.8 and 20.8 points, demonstrating the value of our training corpus for adapting open-source VLMs to temporally grounded, compositional camera motion understanding. Building on these baselines, \inject{} raises micro F1 to 67.5/68.3 on the 4B/8B backbones, gains of 5.4/4.2 points over SFT. \method{} reaches 67.5/67.8, improving over SFT by 5.4/3.7 points. Its macro F1 also increases from 49.0/51.9 to 57.7/57.9, and SegDet@0.5 improves from 33.6/34.9 to 38.2/38.8. These gains support the value of augmenting task supervision with camera-oriented geometric representations, improving both frame-level recognition and complete segment detection. Crucially, \method{} closely matches direct feature injection: both methods achieve 67.5 micro F1 on 4B, and their 8B scores differ by only 0.5 points. The SegDet@0.5 gaps are likewise only 0.4/0.2 points on 4B/8B. Thus, \method{} retains nearly all of the recognition benefit of \inject{} while removing the 3D teacher at inference.

\begin{figure}[!t]
\centering
\includegraphics[width=\linewidth]{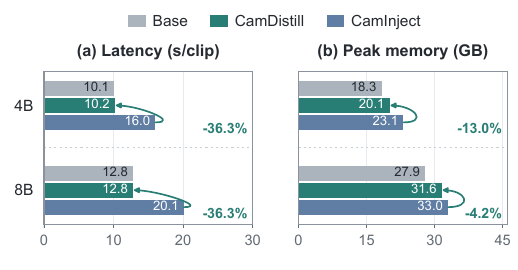}
\caption{\textbf{Inference efficiency.} (a) Latency per clip and (b) peak GPU memory with 4B and 8B backbones. Base and SFT have identical inference costs, so only Base is shown. Measurements use a single H100 GPU with batch size~1. Arrows show percentage reductions achieved by \method{} relative to \inject{}.}
\label{fig:inference_cost}
\end{figure}

\subsection{Analysis and Ablations}
\paragraph{Inference cost.}
Compared with Base (Qwen3-VL), \method{} adds only 0.1\,s/clip on 4B, while 8B latency is unchanged at the reported precision (Figure~\ref{fig:inference_cost}). Peak memory increases by 1.8/3.7\,GB on 4B/8B. Relative to \inject{}, \method{} reduces latency by approximately 36\% on both scales by removing the 3D teacher at inference, while retaining nearly the same recognition performance.

\begin{figure*}[!t]
\centering
\includegraphics[width=\textwidth]{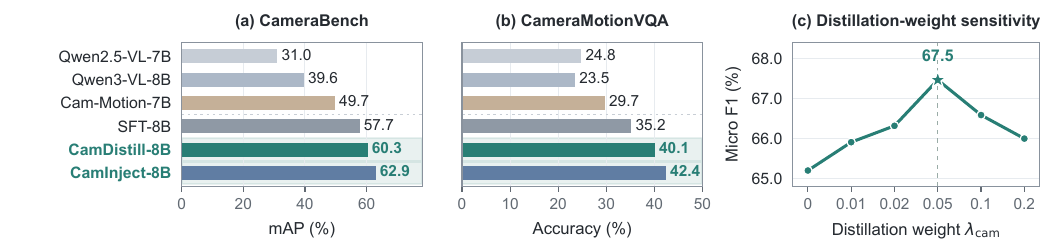}
\caption{\textbf{External generalization and distillation-weight sensitivity.} (a) CameraBench mAP and (b) CameraMotionVQA accuracy. (c) Frame-level micro F1 of \method{}-4B at six tested distillation weights, with the selected $\lambda_{\mathrm{cam}}=0.05$ marked by a star.}
\label{fig:external_sensitivity}
\label{fig:sensitivity}
\end{figure*}

\begin{table}[!t]
\caption{\textbf{Ablations of \method{}-4B.} Each block varies one design factor, namely feature-layer region, module depth, token position, and teacher. Segment metrics use IoU~0.5.}
\label{tab:ablation}
\centering
\footnotesize
\setlength{\tabcolsep}{3pt}
\renewcommand{\arraystretch}{1.17}
\begin{tabular}{@{}L{\dimexpr\linewidth-4cm-8\tabcolsep\relax} *{4}{C{1cm}}@{}}
\toprule
Setting & \shortstack{Micro\\F1} & \shortstack{Macro\\F1} & \shortstack{SegLoc\\@0.5} & \shortstack{SegDet\\@0.5} \\
\midrule
\multicolumn{5}{@{}l}{\textit{Feature-layer region}} \\
Early $(0,3,6,9)$ & 66.1 & 56.0 & 80.0 & 37.3 \\
Late $(14,17,20,23)$ & 65.4 & 54.9 & 79.6 & 36.8 \\
Uniform $(4,9,13,18)$ & 66.8 & 56.5 & 79.9 & 37.6 \\
\rowcolor{oursgreen}
\textbf{Early--middle $(1,5,9,13)$} & \textbf{67.5} & \textbf{57.7} & \textbf{80.4} & \textbf{38.2} \\
\midrule[0.4pt]
\multicolumn{5}{@{}l}{\textit{Module depth}} \\
3 layers $(1,7,13)$ & 66.6 & 56.2 & 79.8 & 37.4 \\
\rowcolor{oursgreen}
\textbf{4 layers $(1,5,9,13)$} & \textbf{67.5} & \textbf{57.7} & \textbf{80.4} & \textbf{38.2} \\
5 layers $(1,4,7,10,13)$ & 67.2 & 57.5 & 80.3 & 38.2 \\
\midrule[0.4pt]
\multicolumn{5}{@{}l}{\textit{Token position}} \\
After visual tokens & 64.7 & 54.5 & 79.5 & 36.9 \\
\rowcolor{oursgreen}
\textbf{Before visual tokens} & \textbf{67.5} & \textbf{57.7} & \textbf{80.4} & \textbf{38.2} \\
\midrule[0.4pt]
\multicolumn{5}{@{}l}{\textit{Teacher}} \\
VGGT & 65.8 & 55.7 & 80.2 & 37.6 \\
\rowcolor{oursgreen}
\textbf{VGGT-$\Omega$} & \textbf{67.5} & \textbf{57.7} & \textbf{80.4} & \textbf{38.2} \\
\bottomrule
\end{tabular}
\end{table}

\paragraph{Generalization to external benchmarks.}
We test the same 8B checkpoints on CameraBench and CameraMotionVQA without additional benchmark-specific training. CameraBench \cite{lin2026towards} evaluates clip-level camera-motion recognition on web videos using mAP; CameraMotionVQA \cite{feng2026geometry} evaluates multiple-choice reasoning on synthetic clips using accuracy. For fair comparison, we follow the official task protocols (Figure~\ref{fig:external_sensitivity}a,b). SFT improves over Base from 39.6 to 57.7 mAP and from 23.5 to 35.2 accuracy. \method{} further reaches 60.3/40.1, gains of 2.6/4.9 over same-data SFT, approaching direct injection at 62.9/42.4. These results demonstrate that models trained on our corpus generalize strongly to other camera motion understanding tasks.

\paragraph{Sensitivity to the distillation weight.}
Figure~\ref{fig:external_sensitivity}c examines the distillation weight on the 4B backbone. At $\lambda_{\mathrm{cam}}=0$, the model retains \gcte{} and camera-token injection but is trained using only the task loss, without geometric supervision. This configuration reaches 65.2 micro F1, compared with 62.1 for SFT, showing that adding the camera-token branch alone improves performance by 3.1 points. Adding geometric supervision with $\lambda_{\mathrm{cam}}=0.05$ further raises micro F1 to 67.5, a gain of 2.3 points with the same branch and trainable parameter count. Thus, the improvement is not solely attributable to the additional \gcte{} parameters: geometric distillation provides a further benefit beyond the branch trained with task supervision alone. Both smaller nonzero weights and larger weights yield lower scores than 0.05, consistent with a balance between task prediction and geometric feature alignment.

\paragraph{Design ablations.}
Table~\ref{tab:ablation} varies one \gcte{} design choice at a time on the 4B backbone. \emph{Feature layers}: early-to-middle features perform best, suggesting that they retain geometric detail while providing sufficient contextual abstraction. Earlier features contain less context, whereas later features are increasingly semantic. The 0.7-point advantage over uniform sampling also favors targeted feature selection. \emph{Module depth}: four alternating blocks achieve the best trade-off. Three blocks trail the default by 0.9 micro F1, while a fifth provides no meaningful gain. \emph{Token position}: placing camera tokens before the visual tokens improves micro F1 by 2.8 points, indicating that they are most useful as conditioning context rather than appended summaries. \emph{Teacher}: our default VGGT-$\Omega$ teacher improves micro/macro F1 over VGGT by 1.7/2.0 points. This suggests that stronger geometric targets improve the student without increasing inference cost. Together with Figure~\ref{fig:sensitivity}, these results attribute the gains to the camera-specific design and supervision rather than to an arbitrary increase in model capacity.

Additional experiments and analyses are provided in Appendices~\ref{app:promptinject}, \ref{app:supervision}, \ref{app:error_analysis}, \ref{app:difficulty}, and~\ref{app:analysis}.

\FloatBarrier
\section{Conclusion}
We study camera motion as a temporally grounded, compositional problem, in which a model must localize when each movement occurs and recover the movements that co-occur within a shot. \dataset{} provides expert interval annotations for this task, and we construct a separate training corpus of approximately 43K clips. The benchmark reveals substantial within-shot variation and concurrent motion. SFT on the training corpus establishes strong baselines. Building on these baselines, we introduce \method{}, which distills the geometry of a 3D foundation model into lightweight camera tokens during training and discards the teacher at inference. It matches the accuracy of direct geometric injection while adding almost no inference cost. We hope \dataset{} and \method{} encourage modeling camera motion as a time-varying, compositional signal.

\clearpage
\section*{LLM Usage Statement}
We used OpenAI Codex to revise author-written text, draft selected passages under author direction, compare related papers, convert and format the LaTeX manuscript, and generate and refine plotting code using author-provided results. Codex also assisted with presenting and interpreting experimental comparisons specified by the authors. The research methods and experiments were developed and conducted by the authors, and all reported experimental results were supplied by the authors. During benchmark curation, a camera motion SFT model provided provisional labels for candidate-pool balancing; the final temporal annotations were produced by human experts. The authors reviewed the AI-assisted content and take full responsibility for the manuscript, code, citations, and scientific claims.

{\small
  \bibliographystyle{ieeenat_fullname}
  \bibliography{main}
}
\clearpage
\onecolumn
\begin{center}
{\Large\bfseries\thetitle\par}
\vspace{0.5em}
{\large Supplementary Material\par}
\vspace{0.5em}
\end{center}
\appendix
\label{supp:start}

\section{Broader Implications and Reproducibility}
\label{app:broader_repro}
\paragraph{Broader implications.}
Camera-motion understanding describes the behavior of the observer, not only the content of the observed scene. Because camera state affects depth cues, visibility, and the interpretation of object motion, camera-aware representations may benefit spatial reasoning, action understanding, video retrieval, and controllable video generation. These applications also require caution. Cinematic labels reflect production conventions and can remain ambiguous across domains, so downstream systems should preserve uncertainty rather than treat every prediction as an objective description of authorial intent.

\paragraph{Reproducibility.}
The training set is disjoint from the independently collected benchmark. The appendices specify the taxonomy, annotation instructions, prompts, frame- and segment-level metrics, training protocol, teacher alignment, and inference-cost protocol. We will publicly release both the benchmark and the full training dataset of approximately 43K clips, together with their annotations, training/evaluation code, and model checkpoints. Video access will be provided through permitted video files, identifiers, or download scripts, subject to the source licenses.

\section{Benchmark Construction Pipeline}
\label{app:pipeline}
\begin{figure}[ht]
\centering
\includegraphics[width=\linewidth]{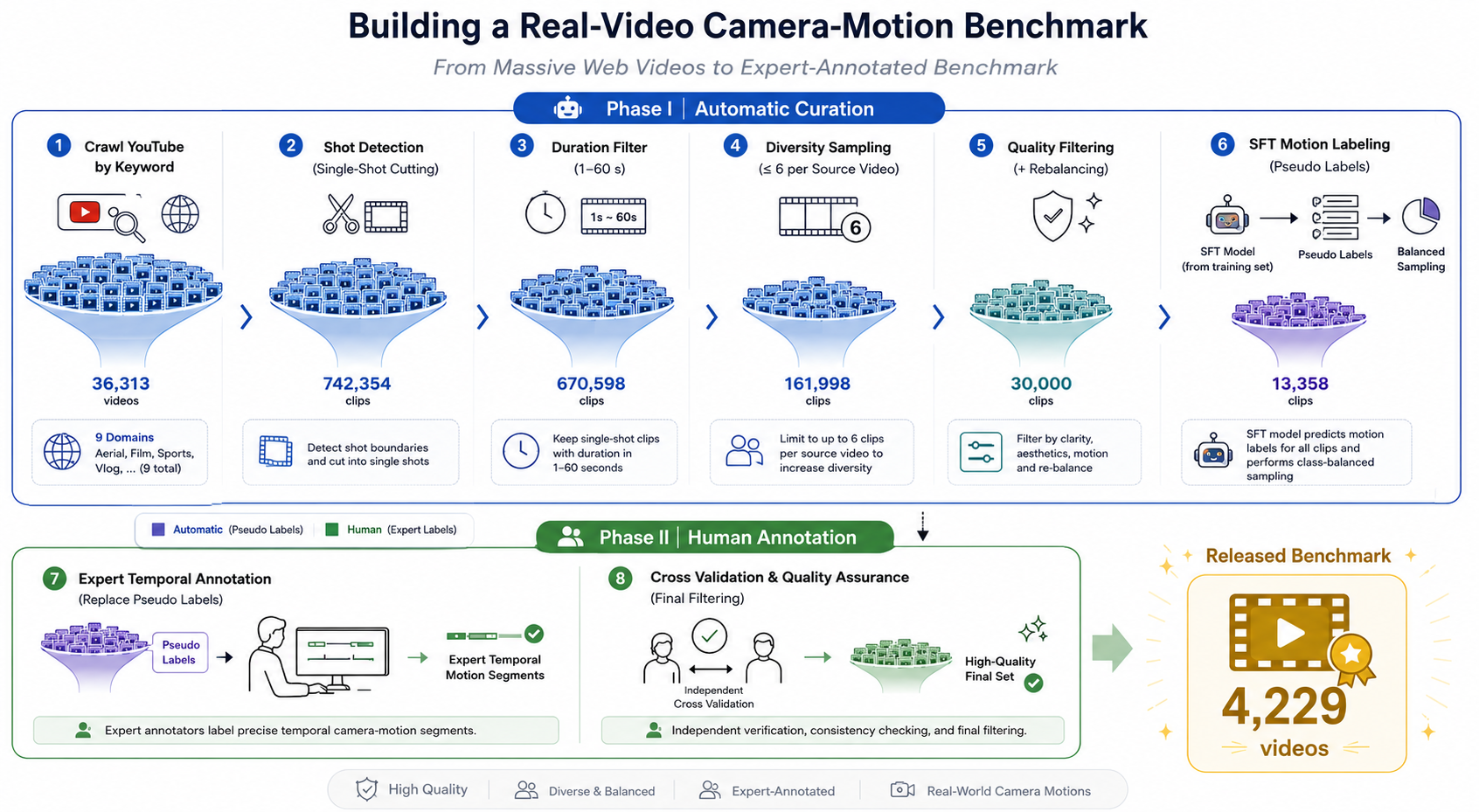}
\caption{\textbf{\dataset{} construction pipeline.} \emph{Phase~I: automatic curation.} We collect YouTube videos from nine domains, split them into shots, filter for duration, diversity, and quality, and use pseudo labels only to balance the candidate pool. This process reduces 36{,}313 source videos to 13{,}358 candidate clips. \emph{Phase~II: human annotation.} Expert annotators replace all pseudo labels with temporal camera-motion annotations. Independent review and final filtering produce 4{,}229 released clips.}
\label{fig:pipeline}
\end{figure}

Figure~\ref{fig:pipeline} summarizes the complete curation pipeline, from source discovery and automatic filtering to class balancing and human quality control.

\textbf{Source collection.} We query YouTube using a curated keyword set that covers all nine content domains, with roughly a dozen queries per domain. Examples include \emph{cinematic footage} and \emph{film clip 4k} for Film~\&~TV; \emph{wildlife documentary} and \emph{aerial nature footage} for Documentary~\&~Nature; \emph{FPV racing drone} and \emph{drone orbit building} for Aerial~\&~Drone; \emph{walking vlog} and \emph{handheld walking footage} for Vlog~\&~Selfie; and \emph{product commercial} and \emph{car advertisement} for Commercial~\&~Ad. We use analogous queries for Sports~\&~Action, Gaming~\&~Animation, Tutorial~\&~Education, and Synthetic~\&~Rendered.

\textbf{Filtering funnel.} Figure~\ref{fig:pipeline} contains eight stages. \textbf{(1)} We retain source videos between 2\,s and 30\,min. \textbf{(2)} TransNetV2 \cite{soucek2024transnet} detects shot boundaries and divides each video into single-shot clips. \textbf{(3)} We keep clips between 1 and 60\,s. \textbf{(4)} At most six clips are retained from each source video to improve content diversity. \textbf{(5)} We re-encode all clips with a uniform \texttt{ffmpeg} profile and score them on clarity, aesthetics, motion intensity, motion quality, and content quality. Clips below threshold on any dimension are removed, while a controlled fraction of near-static clips is restored to preserve the Static prior. \textbf{(6)} An SFT camera-motion model trained on the separate 43{,}438-video training set assigns \emph{pseudo} labels. These labels are used only to construct a balanced candidate pool that up-weights rare movements such as Zoom, Roll, Arc, and Focus Shift. \textbf{(7)} Five annotators with film- and media-related backgrounds independently replace the pseudo labels with precise temporal annotations. \textbf{(8)} Cross-review resolves disagreements by consensus; clips without consensus, as well as doubtful or empty clips, are removed. The resulting benchmark contains 4{,}229 clips.

\section{Taxonomy and Annotation Rules}
\label{app:taxonomy}
\label{app:annotation}
\begin{table}[ht]
\caption{\textbf{Camera motion taxonomy used in \dataset{}.} cw/ccw abbreviate clockwise/counterclockwise.}
\centering
\small
\setlength{\tabcolsep}{6pt}
\renewcommand{\arraystretch}{1.15}
\begin{tabular}{@{}L{2.25cm} L{1.65cm} L{\dimexpr\linewidth-3.90cm-4\tabcolsep\relax}@{}}
\toprule
\textbf{Type} & \textbf{Direction} & \textbf{Annotation cue} \\
\midrule
Static & None & Camera position and orientation remain essentially unchanged; imperceptible micro-jitter is allowed. \\
Unstable & None & Irregular visible shaking without a stable direction; if a stable direction exists, annotate that motion instead. \\
Pan & left/right & Horizontal rotation around a fixed camera position; foreground and background move similarly with little parallax. \\
Tilt & up/down & Vertical rotation around a fixed camera position; frame shifts vertically without translation parallax. \\
Truck & left/right & Lateral camera translation; foreground/background parallax is visible. \\
Pedestal & up/down & Vertical camera translation; perspective and horizon height change. \\
Dolly In/Out & None & Forward/backward camera translation with depth parallax; distinct from zoom. \\
Zoom In/Out & None & Focal-length change with approximately uniform image scaling and no depth parallax. \\
Roll & cw/ccw & Rotation around the optical axis; horizon tilts. \\
Arc & cw/ccw & Camera orbits around a subject or scene center. \\
Follow & None & Camera tracks a moving subject; requires reasoning about subject-centered motion. \\
Focus Shift & None & Focus plane changes while camera motion may be absent. \\
\bottomrule
\end{tabular}
\end{table}

\textbf{Annotation protocol.} The taxonomy contains 12 movement types and 20 direction-aware labels. Directions follow the physical camera motion rather than the apparent background flow: Pan Left/Right denotes rotation of the viewing direction, Truck Left/Right translation of the camera center in its local frame, and Tilt/Pedestal Up/Down the corresponding physical rotation/translation. Each movement also receives a coarse speed attribute (\texttt{zero}, \texttt{slow}, \texttt{medium}, or \texttt{fast}), although speed is not evaluated in this work. Annotators apply seven main rules. (i) Each segment contains all basic movements with perceptible magnitude and clear intent, while minor compensatory motion is ignored. (ii) Direction is constrained by movement type: Pan and Truck use left/right, Tilt and Pedestal use up/down, Arc and Roll use clockwise/counterclockwise, and all remaining types use \texttt{null}. A movement with two directional components is represented by two elements, such as Dolly~In $+$ Tilt~up. (iii) Static permits only imperceptible micro-jitter, while clearly visible directionless shake is labeled Unstable. (iv) Dolly is distinguished from Zoom through depth parallax and perspective change, rather than uniform image scaling. (v) Focus Shift, including rack and follow focus, is treated as a basic movement. (vi) Follow and Arc are annotated together with their underlying primitive, such as Dolly~In, Truck, or Pan. Arc requires the camera to move along a clear curved trajectory around one or more identifiable subjects through at least $45^\circ$; weak curvature below this threshold or motion without a locked subject is treated as a minor adjustment and is not labeled Arc. Arc clockwise/counterclockwise is defined by the camera trajectory around the subject as viewed from above. (vii) Subject motion is not labeled as camera motion. For example, a walking person does not imply camera movement unless the background perspective or frame boundaries also change. Segment boundaries are placed at 0.1-second resolution whenever the active motion phase changes.

\section{Evaluation Protocol Details}
\label{app:metrics}
We formalize the frame- and segment-level metrics summarized in the main text.

\paragraph{Setup.} Each video is represented as a sequence of non-overlapping temporal segments that fully covers $[0,t_{\max}]$, where $t_{\max}$ is the video duration in seconds, and every segment carries a \emph{set} of camera-movement labels. Directional movements are encoded jointly with their direction, for example \texttt{Pan\_left}, while non-directional movements use only the type name. This produces the 20-label space $\mathcal{C}$ used throughout the paper. Metrics are computed on videos present in both the predictions and the ground truth (GT). Missing predicted intervals within an evaluated clip are treated as empty label sets.

\subsection{Frame-Level Metrics}
\paragraph{Sampling.} We sample each timeline at intervals of $\Delta=0.1$\,s. At timestamp $t$, let $Y_t\subseteq\mathcal{C}$ and $\hat{Y}_t\subseteq\mathcal{C}$ denote the GT and predicted label sets of the segments covering $t$, where $s$ and $e$ are its start and end timestamps and it is active when $s\leq t<e$. Because the GT segments cover $[0,t_{\max}]$, evaluation samples the full GT timeline; a missing predicted segment yields an empty predicted label set and therefore false negatives. Because labels are evaluated densely over time, a boundary error affects multiple timestamps and is reflected in the recognition score.

\paragraph{Micro precision/recall/F1.} We accumulate true-positive ($\mathrm{TP}$), false-positive ($\mathrm{FP}$), and false-negative ($\mathrm{FN}$) label counts over all sampled frames; $|\cdot|$ denotes set cardinality here:
\begin{equation}
\mathrm{TP}=\sum_t |Y_t \cap \hat{Y}_t|,\quad
\mathrm{FP}=\sum_t |\hat{Y}_t \setminus Y_t|,\quad
\mathrm{FN}=\sum_t |Y_t \setminus \hat{Y}_t|,
\end{equation}
and define precision $P=\mathrm{TP}/(\mathrm{TP}+\mathrm{FP})$, recall $R=\mathrm{TP}/(\mathrm{TP}+\mathrm{FN})$, and their harmonic mean $\mathrm{F1}=2PR/(P+R)$. Micro averaging is instance-weighted and is therefore dominated by frequent classes.

\paragraph{Macro precision/recall/F1.} We instead compute per-class counts $\mathrm{TP}_c,\mathrm{FP}_c,\mathrm{FN}_c$ for each label $c\in\mathcal{C}$ (a frame contributes to class $c$ as TP if $c\in Y_t\cap\hat{Y}_t$, FP if $c\in\hat{Y}_t\setminus Y_t$, FN if $c\in Y_t\setminus\hat{Y}_t$), form the per-class precision $P_c$, recall $R_c$, and F1 score $\mathrm{F1}_c$, and average them equally over $\mathcal{C}$. Macro averaging is class-weighted and surfaces rare classes such as Roll and Arc. A direction-agnostic variant that collapses each composite label to its type isolates direction errors.

\subsection{Segment-Level Metrics}
\paragraph{Temporal IoU and matching.} This axis operates on whole segments and \emph{decouples} localization from recognition. For a GT segment $g$ and predicted segment $p$, superscripts $\mathrm{start}$ and $\mathrm{end}$ denote their boundary timestamps, while $|g|$, $|p|$, and $|g\cap p|$ denote segment and intersection durations. Their temporal intersection over union (IoU) is
\begin{equation}
\mathrm{IoU}(g,p)=\frac{\max\!\big(0,\ \min(g^{\mathrm{end}},p^{\mathrm{end}})-\max(g^{\mathrm{start}},p^{\mathrm{start}})\big)}{|g|+|p|-|g\cap p|}.
\end{equation}
Given a temporal-IoU threshold $\tau$, we enumerate all GT--prediction segment pairs within each video whose IoU is at least $\tau$, sort them in descending IoU order, and greedily accept a pair only if neither segment has already been matched. This produces a one-to-one, label-agnostic temporal matching. Let $N_G$ and $N_P$ denote the total numbers of GT and predicted segments across all evaluated videos.

\paragraph{Localization (SegLoc).} A predicted segment is a true positive iff it is matched with $\mathrm{IoU}\geq\tau$, \emph{regardless of labels}. With $M_{\mathrm{loc}}$ denoting the total number of matched pairs across all evaluated videos, localization precision $P_{\mathrm{loc}}=M_{\mathrm{loc}}/N_P$, localization recall $R_{\mathrm{loc}}=M_{\mathrm{loc}}/N_G$, and $\text{SegLoc-F1}=2P_{\mathrm{loc}}R_{\mathrm{loc}}/(P_{\mathrm{loc}}+R_{\mathrm{loc}})$. This measures pure temporal segmentation quality.

\paragraph{Detection (SegDet).} Segment detection additionally requires the predicted label set $L_p$ and GT label set $L_g$ to match exactly, $L_p=L_g$. Let $M_{\mathrm{det}}$ denote the total number of matched pairs satisfying this condition across all evaluated videos. We define detection precision $P_{\mathrm{det}}=M_{\mathrm{det}}/N_P$ and recall $R_{\mathrm{det}}=M_{\mathrm{det}}/N_G$, and compute SegDet-F1 analogously. Because $M_{\mathrm{det}}\leq M_{\mathrm{loc}}$, SegDet-F1 cannot exceed SegLoc-F1 at the same threshold. SegDet assigns no partial credit within a segment: a direction error or a missing co-occurring movement invalidates the match. It is therefore substantially stricter than frame-level F1.

\paragraph{Thresholds.} All segment-level metrics are reported separately at $\tau\in\{0.3,0.5,0.7\}$, without averaging across thresholds. Greedy matching is performed independently at each threshold. Ties in IoU are broken by GT segment index and then prediction segment index, making the matching deterministic.

\section{Training and Inference Prompt}
\label{app:prompt}
We use an identical prompt at training and inference. The model receives a video, the system prompt below, and a short user instruction, and is required to return only JSON. The evaluator tolerates minor formatting variations but enforces the closed taxonomy, valid directions, chronological ordering, and non-overlapping temporal segments.

\begin{promptbox}[System Prompt]
\footnotesize
You are a senior film cinematographer. After watching the video, determine which camera movements make up this video, locate their time spans, and output structured JSON.

\textbf{Core principle:} judge only the motion of the camera (lens) itself, not the motion of objects within the frame. People walking or cars driving inside the frame do not mean that the camera is moving. Observe whether the \emph{background and frame edges} move.

\medskip
\textbf{Basic movement} (required, array). Each element is \texttt{\{"type", "direction", "speed"\}}, with \texttt{type} drawn from the closed set below.

\emph{Static / non-steady.}
\texttt{Static}: camera position and orientation essentially unchanged (barely visible micro-jitter allowed); \texttt{speed}=\texttt{zero} if completely still, else \texttt{slow}.
\texttt{Unstable}: clearly perceptible irregular shaking with no stable direction; if a sustained direction exists (e.g.\ the background sweeps left), label that movement instead (e.g.\ Pan).

\emph{Rotation vs.\ translation.}
\texttt{Pan} (fixed camera, horizontal rotation; no depth parallax) vs.\ \texttt{Truck} (lateral camera translation; obvious parallax). \texttt{Tilt} (fixed camera, vertical rotation) vs.\ \texttt{Pedestal} (vertical translation; horizon height and pitch change).

\emph{Depth.}
\texttt{Dolly In}/\texttt{Dolly Out} (camera moves forward/backward; near and far regions scale at different rates because of parallax) vs.\ \texttt{Zoom In}/\texttt{Zoom Out} (focal-length change, camera fixed; uniform scaling, no parallax).

\emph{Other.}
\texttt{Roll} (rotation about the optical axis; tilting horizon), \texttt{Arc} (camera orbits a centered subject), \texttt{Follow} (camera tracks a moving subject; background changes continuously), \texttt{Focus Shift} (focus moves across depth layers). Arc and Follow must also annotate the underlying basic movements (e.g.\ Truck, Pan, Dolly In).

\medskip
\textbf{Direction rules.} \texttt{Truck}$\to$left/right, \texttt{Pedestal}$\to$up/down, \texttt{Pan}$\to$left/right, \texttt{Tilt}$\to$up/down, \texttt{Arc}/\texttt{Roll}$\to$clockwise/counterclockwise; all other types use \texttt{null}.

\textbf{Speed rules.} \texttt{zero} (completely still) / \texttt{slow} (confirmable only on careful inspection) / \texttt{medium} (clearly perceived) / \texttt{fast} (rapid, with a sense of speed).

\textbf{Compound movement.} When several motions occur simultaneously and are all observable, output all items (e.g.\ Dolly In while Pan). Output the \texttt{segments} array ordered by time, following the format below, and output \emph{only} JSON.
\end{promptbox}

\begin{codebox}[Output Format: Example JSON]
\footnotesize
\begin{verbatim}
{
  "segments": [
    {
      "start_time": 0.0,
      "end_time": 2.5,
      "basic_movement": [
        {"type": "Truck", "direction": "right", "speed": "medium"},
        {"type": "Dolly In", "direction": null, "speed": "slow"}
      ],
      "confidence": "high"
    },
    {
      "start_time": 2.5,
      "end_time": 5.0,
      "basic_movement": [
        {"type": "Static", "direction": null, "speed": "zero"}
      ],
      "confidence": "high"
    }
  ]
}
\end{verbatim}
\end{codebox}

\begin{promptbox}[User Prompt]
\footnotesize
Analyze the camera movement in this video and output JSON following the system prompt rules. Output only JSON.
\end{promptbox}

\section{Additional Implementation Details}
\label{app:implementation}
The camera module uses QK-normalized scaled dot-product attention, pre-normalization, residual connections, and small LayerScale coefficients. The first-frame and shared subsequent-frame queries are initialized from a zero-mean Gaussian with standard deviation $10^{-3}$. For each video, the teacher cache stores an $S \times 2048$ tensor, where $S$ is the number of teacher frames. Before computing the loss, these features are aligned to the VLM temporal grid. At inference, \inject{} runs the frozen 3D teacher online to extract camera tokens and maps them to the LLM hidden size with a two-layer projector of approximately 8M parameters. \method{} accesses the cache only during training.

\textbf{Training data.} SFT, \method{}, and \inject{} are trained on our corpus of approximately 43K clips (Section~\ref{sec:training_corpus}). The training annotations use the same 12 movement types, 20 direction-aware labels, 0.1-second temporal resolution, and annotation protocol as \dataset{} (Appendix~\ref{app:annotation}). The data source is the main difference: the training clips come from Tencent Video, whereas the benchmark clips come from YouTube. The two sets are disjoint. We additionally screen them with CLIP embeddings and manually inspect high-similarity pairs; no duplicate clip is retained.

\textbf{Evaluation inputs.} Qwen-family open-source models, Cam-Motion-7B, SFT, \method{}, and \inject{} use 5 FPS with at most 100 frames. They share the same English prompt, deterministic decoding, parser, and evaluator. Gemini receives the video through its video interface with a requested 5-FPS rate. GPT-5.4 receives uniformly sampled timestamped frames (5 FPS) in the evaluated API configuration. InternVL3.5 uses 32 uniformly sampled frames because its inference interface accepts a fixed frame count. External CameraBench and CameraMotionVQA results use the benchmarks' official scripts and protocols without our temporal JSON prompt.

\textbf{Training configuration.} Table~\ref{tab:train_config} gives the complete setup for both backbones. We fully fine-tune the language model while freezing the vision encoder and visual projector. We train \gcte{} jointly with the language model. For \inject{}, the teacher projector is also trainable. \gcte{} adds approximately 110.2M trainable parameters to the 4B backbone and 216.3M to the 8B backbone. Block $m$ attends to the $m$-th selected ViT layer, so module depth equals the number of tapped layers.

When the number of teacher frames differs from the number of VLM temporal groups, we align the teacher sequence using two-frame average pooling when possible, followed by adaptive pooling or nearest-neighbor interpolation when necessary. Because the 3D model provides only supervision for \method{}, we extract its camera tokens once and cache them offline. The teacher is not invoked in the optimization loop, so distillation adds little runtime overhead beyond standard SFT. For \inject{}, the teacher can alternatively be run online for each batch.

\begin{table}[ht]
\caption{\textbf{Training hyperparameters.} Shared by \method{} and \inject{} unless a per-backbone value is given.}
\label{tab:train_config}
\centering
\resizebox{0.82\linewidth}{!}{
\begin{tabular}{L{6.2cm} L{7.0cm}}
\toprule
Setting & Value \\
\midrule
\rowcolor{headerblue}\multicolumn{2}{l}{\emph{Model and optimization}} \\
Backbone & Qwen3-VL-4B / 8B-Instruct \\
Trainable parameters & full LLM; \gcte{} module \\
Frozen parameters & vision encoder, visual projector \\
Precision / attention & bfloat16 / FlashAttention-2 \\
Memory & gradient checkpointing, DeepSpeed ZeRO-2 \\
GPUs & $8\times$ H20 \\
Per-device batch size & 2 \\
Gradient accumulation & 4 (effective batch size 64) \\
Epochs & 2 \\
Learning rate (4B / 8B) & $2{\times}10^{-5}$ / $1.5{\times}10^{-5}$ \\
LR schedule & cosine, warm-up ratio 0.05 \\
Weight decay & 0.01 \\
Max sequence length & 16{,}384 \\
Training time (4B / 8B) & ${\sim}7$ / ${\sim}10$ hours \\
\midrule
\rowcolor{headerblue}\multicolumn{2}{l}{\emph{Video sampling}} \\
Frame rate / count & 5 FPS, 4--100 frames \\
Max pixels per frame & 100{,}352 \\
\midrule
\rowcolor{headerblue}\multicolumn{2}{l}{\emph{Camera-token module and distillation}} \\
\gcte{} depth $M$ (4B / 8B) & 4 / 8 \\
Tapped ViT layers (4B) & $\{1,5,9,13\}$ over 24 layers \\
Tapped ViT layers (8B) & $\{1,3,5,7,9,11,13,15\}$ \\
Camera-token dim $d_c$ & 1024 (concatenated feature 2048) \\
Distillation weight $\lambda_{\mathrm{cam}}$ & 0.05 \\
Teacher & 3D foundation model (VGGT / VGGT-$\Omega$), cached \\
\bottomrule
\end{tabular}}
\end{table}

\begin{figure}[ht]
\centering
\includegraphics[width=\linewidth]{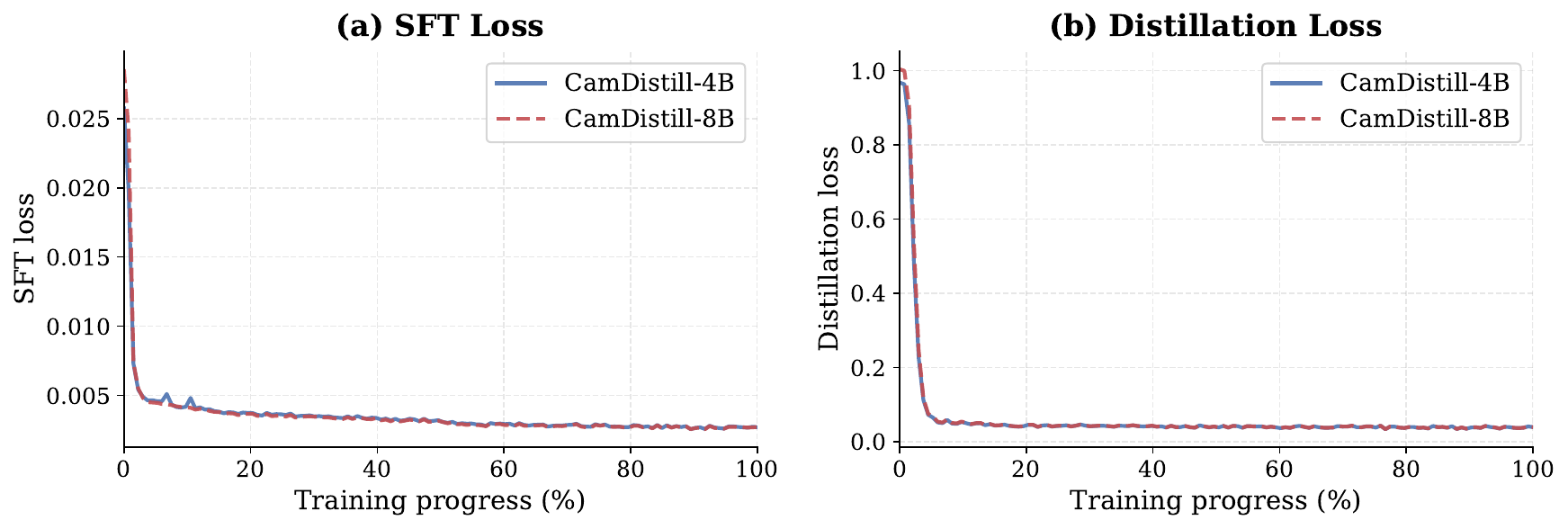}
\caption{\textbf{Training loss curves.} \textbf{(a)} Next-token SFT loss and \textbf{(b)} camera-token distillation loss for \method{}-4B (solid) and \method{}-8B (dashed) over normalized training progress. Both backbones converge to similar final values (SFT $\approx0.003$, distillation $\approx0.04$), indicating stable optimization across model scales. Curves are logged directly without additional smoothing.}
\label{fig:loss_curves}
\end{figure}

\textbf{Training curves.} Figure~\ref{fig:loss_curves} reports the SFT and distillation losses for both backbones. Both objectives decrease smoothly, and the 4B and 8B curves nearly overlap. The training-set cosine-distance loss falls from approximately $1$ to $0.04$, showing that \gcte{} fits the cached teacher targets during optimization.

\section{Camera Pose Estimation Methods and Teacher Choice}
\label{app:pose}
Our camera tokens are distilled from a 3D foundation model that estimates camera geometry. We briefly review candidate estimators and explain our teacher choice.

Classical structure-from-motion and SLAM systems, such as COLMAP \cite{schonberger2016structure} and DROID-SLAM \cite{teed2021droid}, recover trajectories through feature matching and geometric optimization. These pipelines can be slow and brittle under low texture or pure rotation. They are also sensitive to dynamic subjects, which are common in film, television, and vlog footage: when a moving person occupies much of the frame, feature matching may attribute subject motion to the camera.

Feed-forward geometry transformers instead learn scene-level priors. DUSt3R \cite{wang2024dust3r}, VGGT \cite{wang2025vggt}, and VGGT-$\Omega$ \cite{wang2026vggt} jointly estimate camera parameters, depth, and point maps in a single forward pass without bundle adjustment. Permutation-equivariant models such as $\pi^3$ \cite{wang2026pi} improve multi-view consistency and long-sequence stability. ViPE \cite{huang2025vipe} jointly estimates depth and pose in low-texture and high-motion scenes, Reloc3r \cite{dong2025reloc3r} focuses on relative-pose regression, and DROID-SLAM in the Wild \cite{li2026droid} improves SLAM robustness in dynamic environments. Large annotated resources such as SpatialVID \cite{wang2025spatialvid} further support progress in video geometry.

We select \textbf{VGGT / VGGT-$\Omega$} as teachers for three reasons. First, their feed-forward inference avoids per-scene optimization. Second, their joint reasoning over cameras and 3D structure is better suited to shots dominated by dynamic subjects than matching-based SfM. Third, they produce one compact 2048-dimensional camera token per frame, providing a direct geometry-aligned target for \gcte{}.

\section{Geometry-Only Baseline Details}
\label{app:geobaseline}
The geometry-only baseline (\S\ref{sec:experiments}) converts VGGT-$\Omega$ camera poses into \dataset{}-style segment labels using the same deterministic pipeline for all benchmark clips. From consecutive poses it computes camera-local translation $(\Delta x,\Delta y,\Delta z)$, local yaw/pitch/roll, world speed, and trajectory curvature. These eight signals are z-normalized for segmentation.

\textbf{Segmentation.} Frames below both a translation-speed threshold of $0.005$ and an angular-speed threshold of $0.3^\circ$ are grouped into static intervals of at least three frames. Their boundaries are fixed, and the remaining intervals are segmented with PELT change-point detection using an RBF cost, penalty 3.0, and minimum segment length three. Segments shorter than three frames are merged into the longer neighbor, and adjacent segments with identical label sets are merged.

\textbf{Threshold classification.} Within each segment, local translation and angular velocities are averaged and each axis is thresholded independently. The evaluated pipeline uses a translation threshold $\tau_t{=}0.02$ in the pose encoder's local units and a rotation threshold $\tau_r{=}0.5^\circ$:
\begin{center}
\small
\begin{tabular}{l l}
\toprule
Condition & Label \\
\midrule
$\Delta z > \tau_t$ / $<-\tau_t$ & Dolly In / Dolly Out \\
$\Delta x > \tau_t$ / $<-\tau_t$ & Truck Right / Truck Left \\
$\Delta y > \tau_t$ / $<-\tau_t$ & Pedestal Down / Pedestal Up \\
$\text{yaw} > \tau_r$ / $<-\tau_r$ & Pan Left / Pan Right \\
$\text{pitch} > \tau_r$ / $<-\tau_r$ & Tilt Down / Tilt Up \\
$\text{roll} > \tau_r$ / $<-\tau_r$ & Roll CCW / Roll CW \\
no axis exceeds its threshold & Static \\
\bottomrule
\end{tabular}
\end{center}
Multiple axes can exceed their thresholds simultaneously, producing compound camera motion. Arc is approximated when lateral translation and yaw have absolute Pearson correlation at least 0.7 with sufficient amplitude; Follow requires sustained forward motion and moderate trajectory curvature. Unstable is detected at the video level when the high-frequency FFT energy ratio of raw lateral motion exceeds 0.3, using 0.2 of the spectrum as the low-frequency cutoff and at least eight frames. The baseline cannot predict Zoom, which depends on focal-length change rather than extrinsics, or Focus Shift, which is optical rather than geometric. Its hand-designed rules are intended as an interpretable pose-only reference, not a competitive learned model.

\section{Pose-as-Text Prompting on 4B and 8B}
\label{app:promptinject}
Table~\ref{tab:promptinject} compares PromptInject with its untuned base model on both backbone scales. Adding teacher-estimated poses as text improves segment localization from 55.6 to 67.3 on 4B and from 67.7 to 74.5 on 8B, while micro F1 changes from 24.2 to 23.1 and from 28.3 to 28.1, respectively. Reported latency increases from 10.1 to 16.8 seconds per clip on 4B and from 12.8 to 21.5 on 8B; peak memory also increases. These results characterize pose-as-text prompting; the trained SFT and camera-token models are compared in Table~\ref{tab:main_results}, with their costs shown in Figure~\ref{fig:inference_cost}.

\begin{table}[ht]
\centering
\caption{\textbf{Pose-as-text prompting on 4B and 8B.} PromptInject adds teacher-estimated poses to the base model's prompt. Frame-level micro/macro F1 and segment-level F1 at IoU~0.5 use the same evaluation protocol. Reported latency and peak memory use a single H100 at batch~1.}
\label{tab:promptinject}
\small
\setlength{\tabcolsep}{8pt}
\renewcommand{\arraystretch}{1.15}
\begin{tabular}{@{}llrrrrrr@{}}
\toprule
Backbone & Method & Micro F1 & Macro F1 & SegLoc@0.5 & SegDet@0.5 & \shortstack{Latency\,(s/clip)\\$\downarrow$} & \shortstack{Memory\,(GB)\\$\downarrow$} \\
\midrule
\multirow{2}{*}{4B} & Qwen3-VL & 24.2 & 7.8 & 55.6 & 11.6 & 10.1 & 18.3 \\
 & +PromptInject & 23.1 & 10.7 & 67.3 & 12.1 & 16.8 & 24.0 \\
\midrule
\multirow{2}{*}{8B} & Qwen3-VL & 28.3 & 9.1 & 67.7 & 15.0 & 12.8 & 27.9 \\
 & +PromptInject & 28.1 & 12.2 & 74.5 & 18.6 & 21.5 & 33.7 \\
\bottomrule
\end{tabular}
\end{table}

\FloatBarrier

\section{Structure and Auxiliary-Supervision Ablation}
\label{app:supervision}
Table~\ref{tab:supervision} compares four configurations on the 4B backbone. All configurations use the same structured camera motion task loss and freeze the vision encoder while fine-tuning the language model. SFT uses no camera-token branch. The branch-only configuration adds \gcte{} and camera-token injection without geometric supervision. \method{} retains the same branch and projector and adds camera-token feature distillation. The pose-supervised configuration uses the pose-decoding head and pose loss from VGGT-$\Omega$ \cite{wang2026vggt}, with target poses estimated by the same frozen VGGT-$\Omega$ teacher. This replaces camera-token feature distillation with pose supervision, and the pose head is discarded at inference.

\begin{table}[!ht]
\caption{\textbf{Structure and auxiliary-supervision ablation on the 4B backbone.} All configurations retain supervised fine-tuning with the structured camera-motion task loss. B--D use the Geometry-aware Camera Token Extractor; B and C share its branch and projector parameters. D uses the VGGT-$\Omega$ pose-decoding head during training and removes it at inference. Micro and macro F1 evaluate frame-level labels; SegLoc and SegDet use temporal IoU~0.5.}
\label{tab:supervision}
\centering
\small
\setlength{\tabcolsep}{4pt}
\renewcommand{\arraystretch}{1.25}
\begin{tabular}{@{}C{0.5cm} L{\dimexpr\linewidth-8.7cm-12\tabcolsep\relax} L{3.6cm} *{4}{C{1.15cm}}@{}}
\toprule
ID & Model configuration & Auxiliary supervision & \shortstack{Micro\\F1} & \shortstack{Macro\\F1} & \shortstack{SegLoc\\@0.5} & \shortstack{SegDet\\@0.5} \\
\midrule
A & Supervised fine-tuning of the language model & None & 62.1 & 49.0 & 78.9 & 33.6 \\
\midrule[0.4pt]
B & Language model + Geometry-aware Camera Token Extractor & None (task loss only) & 65.2 & 53.7 & 79.5 & 36.1 \\
\rowcolor{oursgreen}
\textbf{C} & \textbf{Language model + Geometry-aware Camera Token Extractor (\method{})} & \textbf{Camera-token feature distillation} & \textbf{67.5} & \textbf{57.7} & \textbf{80.4} & \textbf{38.2} \\
D & Language model + Geometry-aware Camera Token Extractor + VGGT-$\Omega$ pose head (training only) & Teacher-estimated camera poses & 66.5 & 54.1 & 80.3 & 38.1 \\
\bottomrule
\end{tabular}
\end{table}

The branch-only configuration corresponds to $\lambda_{\mathrm{cam}}=0$ in Figure~\ref{fig:external_sensitivity}c and reaches 65.2 micro F1, compared with 62.1 for SFT. Adding geometric feature distillation further improves micro F1 to 67.5 with the same branch and trainable parameter count. Pose supervision reaches 66.5 micro F1 and 54.1 macro F1, while camera-token feature distillation achieves 67.5 and 57.7, respectively. Feature supervision therefore performs better than explicit pose supervision in this comparison. One possible explanation is that poses estimated by VGGT-$\Omega$ contain errors, introducing noise into the pose supervision targets.
\FloatBarrier

\section{Detailed \gcte{} Block Structure}
\label{app:gcte}
\gcte{} alternates two attention operations within each of its $M$ blocks. A \emph{frame-wise cross-attention} lets each frame's camera token query the frozen visual tokens of that frame, and a \emph{global camera self-attention} then lets the per-frame camera tokens exchange information within the video. Both are standard pre-norm transformer sublayers with LayerNorm ($\mathrm{LN}$), QK-normalized attention, LayerScale-gated residuals, and a feed-forward layer, and both write only to the camera tokens, so the pretrained visual stream is unchanged. This section expands the abstract $\mathrm{CrossAttn}$ and $\mathrm{SelfAttn}$ maps of Section~\ref{sec:method} into their sublayer form.

\paragraph{Frame-wise cross-attention.} In block $m$, the camera token $c_i^{(m-1)}$ of frame $i$ queries that frame's frozen features $x_i^{(\ell_m)}$ at the tapped vision layer $\ell_m$. With multi-head cross-attention $\mathrm{MHCA}$ (query from the camera token, keys and values from the frame features) and LayerScale vectors $\gamma_1,\gamma_2$, let $u_i^{(m)}$ denote the camera state after the attention residual update. Here $\mathrm{FFN}$ denotes the feed-forward network and $\odot$ elementwise multiplication:
\begin{align}
u_i^{(m)} &= c_i^{(m-1)} + \gamma_1 \odot \mathrm{MHCA}\big(\mathrm{LN}(c_i^{(m-1)}),\ \mathrm{LN}(x_i^{(\ell_m)})\big), \\
\tilde{c}_i^{(m)} &= u_i^{(m)} + \gamma_2 \odot \mathrm{FFN}\big(\mathrm{LN}(u_i^{(m)})\big).
\end{align}
The keys and values are read-only, so the visual tokens $x_i^{(\ell_m)}$ are never updated.

\paragraph{Global camera self-attention.} The post-cross-attention tokens of a video then attend to one another through multi-head self-attention $\mathrm{MHSA}$, with LayerScale vectors $\gamma_3,\gamma_4$. Let $v_i^{(m)}$ be the camera state after the self-attention residual update, and $\tilde{c}_{1:T}^{(m)}$ the sequence of all post-cross-attention camera tokens:
\begin{align}
v_i^{(m)} &= \tilde{c}_i^{(m)} + \gamma_3 \odot \mathrm{MHSA}\big(\mathrm{LN}(\tilde{c}_{1:T}^{(m)})\big)_i, \\
c_i^{(m)} &= v_i^{(m)} + \gamma_4 \odot \mathrm{FFN}\big(\mathrm{LN}(v_i^{(m)})\big).
\end{align}
The attention is masked block-diagonally, so camera tokens from different videos in a batch do not interact. After the final block, the frame-level and temporally contextualized states are concatenated into the token distilled against the teacher,
\begin{equation}
z_i = \big[\tilde{c}_i^{(M)};\ c_i^{(M)}\big] \in \mathbb{R}^{2 d_c}.
\end{equation}
Algorithm~\ref{alg:gcte} summarizes the full forward pass, using $Q$, $K$, and $V$ for attention queries, keys, and values. Superscripts $\mathrm{frame}$ and $\mathrm{global}$ identify the final cross-attention and self-attention states, respectively.

\begin{algorithm}[ht]
\caption{Forward pass of \gcte{}.}
\label{alg:gcte}
\begin{algorithmic}[1]
\STATE Initialize $c_1$ with the first-frame camera query and $c_{2:T}$ with the shared non-first-frame camera query.
\FOR{$m=1$ to $M$}
    \STATE Select frozen vision feature layer $\ell_m$.
    \FOR{each frame $i$}
        \STATE \textbf{\fcva{}:} update $c_i$ by cross-attention with $Q=c_i$ and $K,V=x_i^{(\ell_m)}$; keep $x_i^{(\ell_m)}$ unchanged.
    \ENDFOR
    \STATE Save the post-\fcva{} tokens in the last block as the frame-level branch.
    \STATE \textbf{\gcsa{}:} for each video independently, update $c_{1:T}$ by self-attention over camera tokens only.
\ENDFOR
\STATE Concatenate the final post-\fcva{} and post-\gcsa{} tokens to obtain $z_i=[c_i^{\mathrm{frame}};c_i^{\mathrm{global}}]$.
\STATE Project $z_i$ to the LLM hidden size and insert it before frame $i$'s visual tokens.
\end{algorithmic}
\end{algorithm}

\paragraph{Implementation notes.} Each camera branch has dimension $d_c=1024$. Concatenating the final block's post-cross-attention and post-self-attention states gives the student token $z_i$, which matches the cached teacher token. The two camera queries are initialized from zero-mean Gaussians (standard deviation $\sigma=10^{-3}$) and all linear layers with Xavier initialization. During training the teacher token supervises $z_i$ through the distillation loss, and at inference the teacher branch is removed so that only the projected $z_i$ enters the LLM.

\paragraph{Position encoding.} The decoder uses multimodal RoPE (M-RoPE), which assigns each token a $(t,h,w)$ coordinate, with $t$ denoting the temporal index and $h,w$ the vertical and horizontal spatial indices. A camera token receives the temporal index of its frame and the spatial center of that frame's visual patch grid. The decoder therefore interprets it as part of the corresponding frame rather than as an additional time step. Because its coordinate is fixed, prepending or appending the token changes sequence order but not its M-RoPE position.

\section{Computational Complexity}
\label{app:complexity}
We analyze only the inference overhead introduced by \method{}, because its 3D teacher is absent at test time. Let $T$ be the number of frames, $P$ the number of frozen visual tokens per frame, $d_v$ the vision width, $d_c$ the camera-token width, and $M$ the number of \gcte{} blocks. We use $O(\cdot)$ to denote asymptotic operation counts.

\paragraph{\gcte{} blocks.}
In each block, frame-wise cross-attention projects the $TP$ frozen visual tokens to keys and values and lets one camera query per frame attend to its $P$ visual tokens. Including projections, attention interactions, and the camera-token feed-forward update, its cost is
\begin{equation}
O\!\left(TP d_v d_c + TP d_c + T d_c^2\right).
\end{equation}
Global camera self-attention operates on only the $T$ camera tokens. Its projections, pairwise attention, and feed-forward update cost
\begin{equation}
O\!\left(T d_c^2 + T^2 d_c\right).
\end{equation}
Across $M$ blocks, \gcte{} therefore adds
\begin{equation}
O\!\left(M\left(TP d_v d_c + TP d_c + T d_c^2 + T^2 d_c\right)\right).
\end{equation}
For fixed hidden widths, this overhead is linear in the number of visual tokens $TP$, apart from self-attention over the much shorter sequence of $T$ camera tokens. The expression includes the visual key/value projections omitted by an attention-interaction-only analysis.

\paragraph{LLM sequence overhead.}
\method{} inserts exactly one camera token per frame. Let $L$ be the number of text tokens; the LLM sequence length grows from $TP+L$ to $TP+T+L$. Relative to the visual sequence, the token-count increase is $T/(TP)=1/P$. This does not make decoder attention free, but it is much smaller than adding dense geometry tokens for every visual patch. The measured system-level effect is reported in Figure~\ref{fig:inference_cost}: latency changes from 10.1 to 10.2 seconds per clip on the 4B backbone, while peak memory increases from 18.3 to 20.1 GB. We therefore describe \method{} as having negligible measured latency overhead, with a modest memory increase, rather than as cost-free.

\paragraph{Summary.}
\method{} is efficient for two reasons. First, \gcte{} replaces the teacher's global attention over all visual tokens with single-query cross-attention and self-attention over only $T$ camera tokens. Second, it increases the LLM sequence length by only a fraction $1/P$. Once the teacher is removed, the resulting inference cost is close to that of the SFT backbone, consistent with the latency and memory measurements in Figure~\ref{fig:inference_cost}.

\section{Compositionality and Error Analysis}
\label{app:error_analysis}
\begin{figure}[t]
\centering
\includegraphics[width=0.92\linewidth]{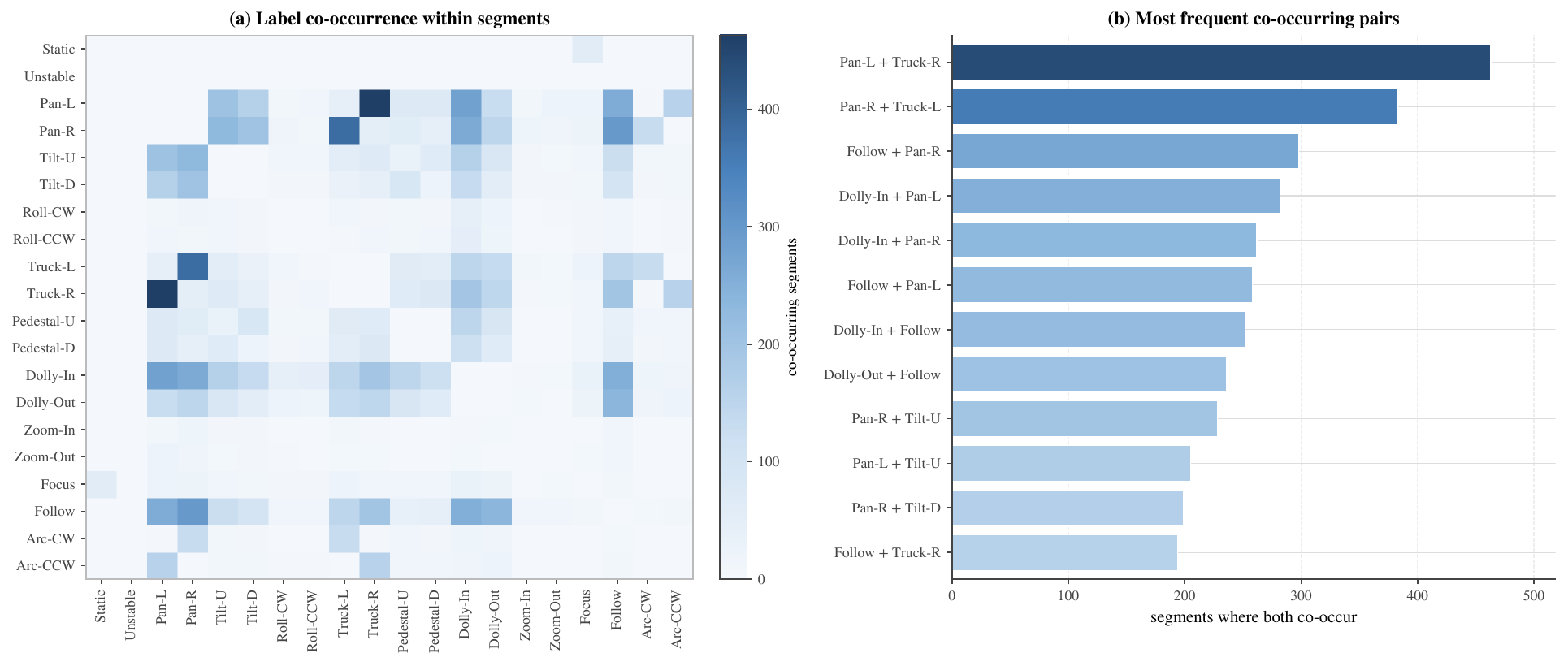}
\caption{\textbf{Camera-motion co-occurrence in \dataset{}.} Left: direction-aware co-occurrence matrix. Right: most frequent label pairs.}
\label{fig:cooc}
\end{figure}

\paragraph{Co-occurrence.}
Camera motion in \dataset{} is strongly compositional: 44.2\% of segments contain compound camera motion with at least two simultaneous movement primitives. Figure~\ref{fig:cooc} reveals recurring structures rather than arbitrary combinations. Pan and Truck often occur in opposite directions during reveals or orbiting shots, Follow commonly accompanies Pan or Dolly, and Pan$+$Tilt represents coordinated rotation. The unusual opposing Dolly--Zoom combination appears in eight segments.

\begin{figure}[t]
\centering
\includegraphics[width=0.92\linewidth]{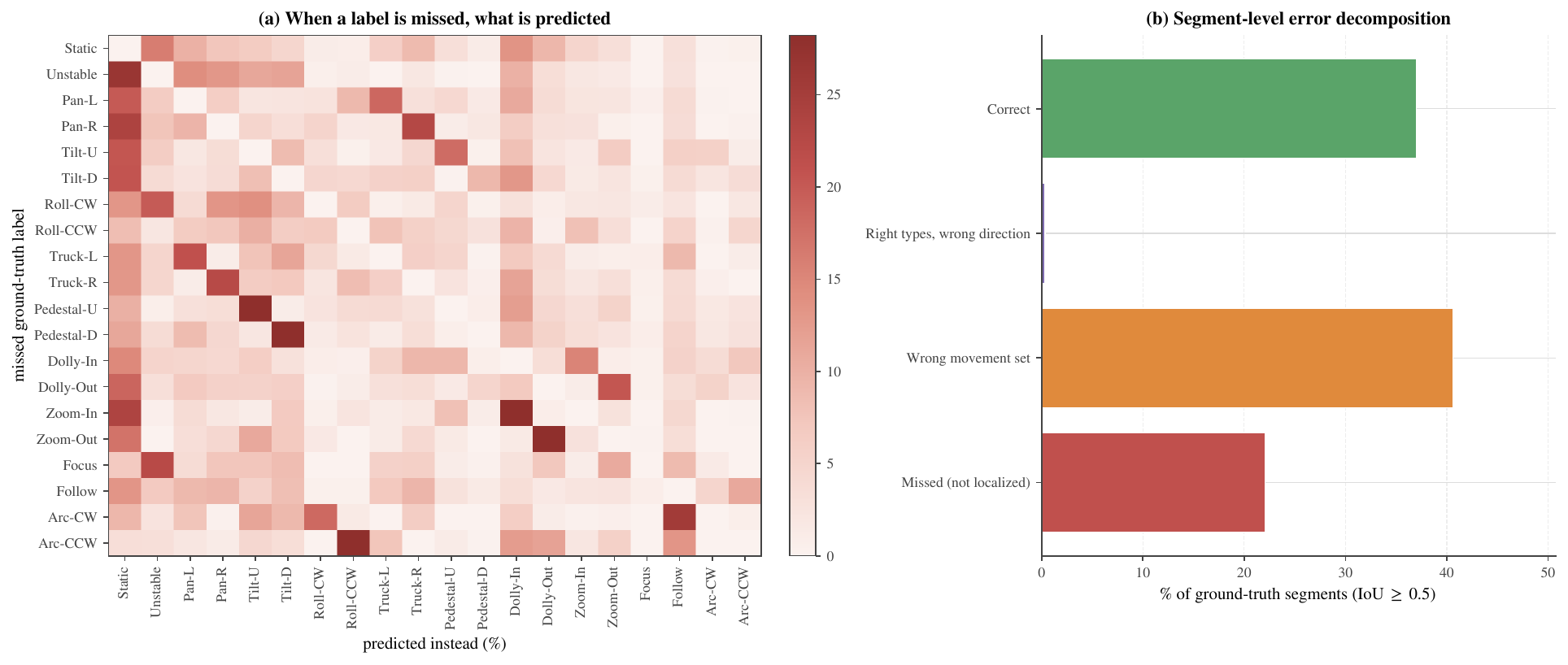}
\caption{\textbf{Error analysis of \method{}-8B.} Left: row-normalized frame-level confusions. Right: segment-level error decomposition after IoU~$\geq0.5$ matching.}
\label{fig:errors}
\end{figure}

\paragraph{Error decomposition.}
At temporal IoU~$\geq0.5$ (Figure~\ref{fig:errors}), \method{}-8B recovers a substantial fraction of ground-truth segments with the exact label set. The largest error category consists of correctly localized segments with incomplete or incorrect movement sets, followed by segments that are not localized. Cases with correct movement types but incorrect directions are rare. The remaining confusions are physically plausible: distinguishing Dolly from Zoom and Pan from Truck requires parallax or lens evidence that is not apparent from image-plane motion alone.

\section{Per-Class Difficulty, Complexity, and Temporal Precision}
\label{app:difficulty}
We next use \method{}-8B predictions to identify where the difficulty of \dataset{} is concentrated.

\begin{figure}[t]
\centering
\includegraphics[width=\linewidth]{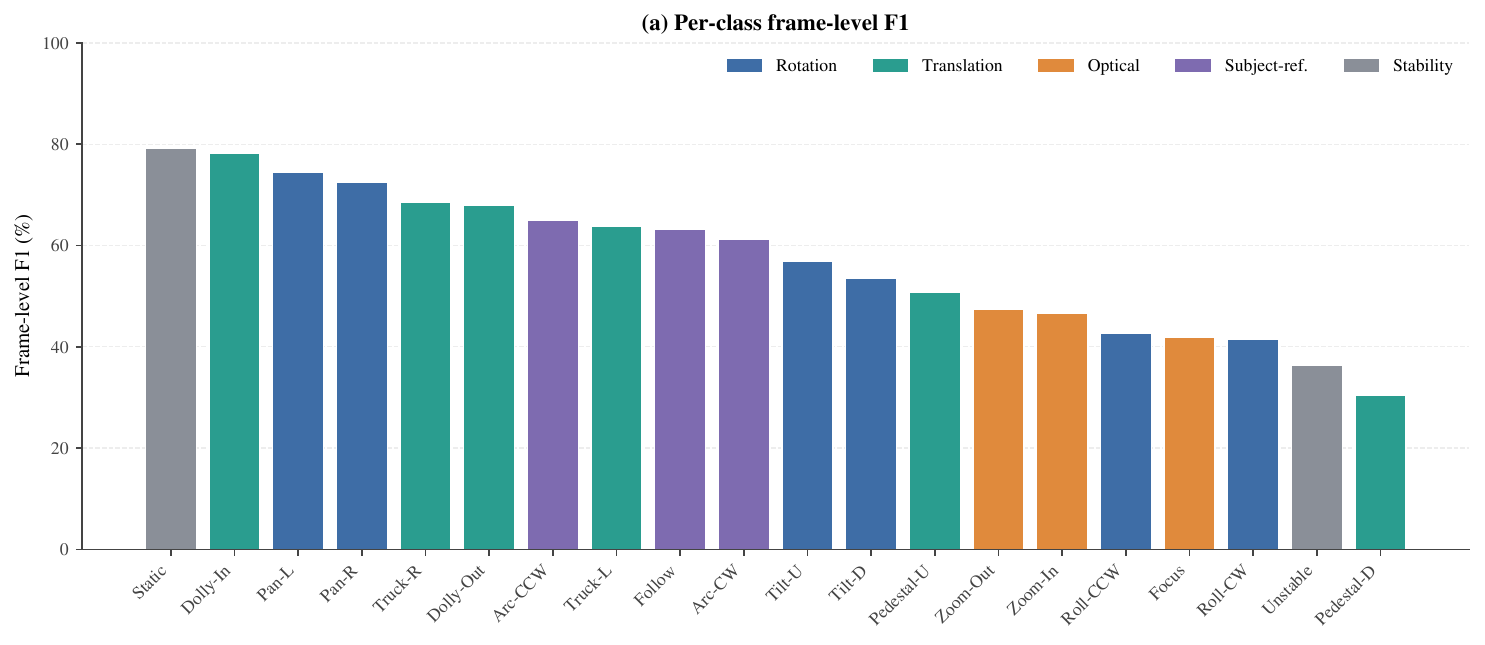}
\caption{\textbf{Per-class frame-level F1 of \method{}-8B}, sorted by score and colored by motion family.}
\label{fig:perclass}
\end{figure}

\paragraph{Per-class difficulty and the long tail.}
Figure~\ref{fig:perclass} reports frame-level F1 for all 20 direction-aware labels. Performance is uneven and broadly follows class frequency: the more frequent labels achieve higher average F1 than the rarer labels. Common and geometrically salient movements such as Static, Dolly, Pan, and Truck are recognized reliably. Rare optical and rotational movements such as Zoom, Roll, and Focus Shift remain difficult because they have fewer training examples and depend on subtle cues rather than large image-plane displacement. Arc, although equally rare, is recognized more reliably, since its curved orbiting trajectory produces distinctive image-plane motion. This class imbalance motivates reporting macro averages alongside micro averages.

\begin{figure}[t]
\centering
\includegraphics[width=\linewidth]{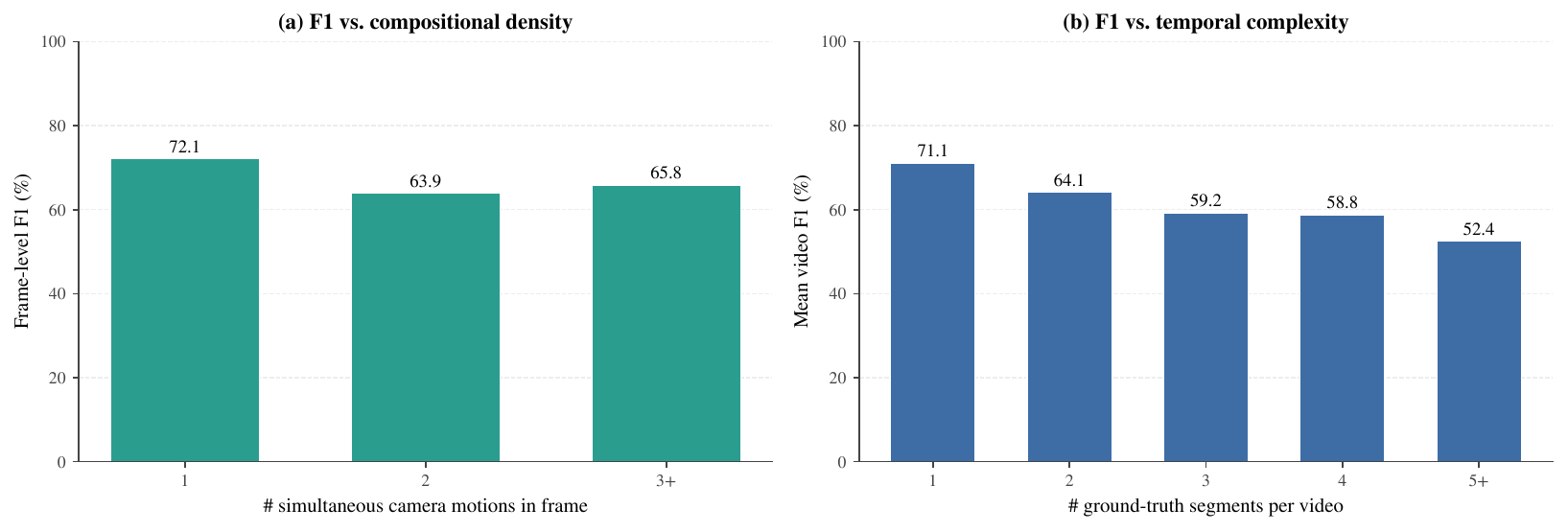}
\caption{\textbf{Performance versus task complexity.} (a) Frame-level F1 decreases as more camera motions co-occur; (b) mean per-video F1 decreases as the clip is split into more temporal segments.}
\label{fig:complexity}
\end{figure}

\paragraph{Difficulty grows with composition and temporal structure.}
Figure~\ref{fig:complexity} isolates the two properties central to our task. In panel (a), frame-level F1 is 72.1 with one active movement, compared with 63.9 for two and 65.8 for three or more, showing that compound frames are harder than single-motion frames without implying a monotonic trend within the compound groups. In panel (b), mean per-video F1 decreases monotonically from 71.1 for clips with one segment to 52.4 for clips with five or more. These trends confirm that both temporal structure and motion composition contribute substantially to the difficulty of \dataset{}.

\begin{figure}[t]
\centering
\begin{minipage}[c]{0.52\textwidth}
\centering
\includegraphics[width=\linewidth]{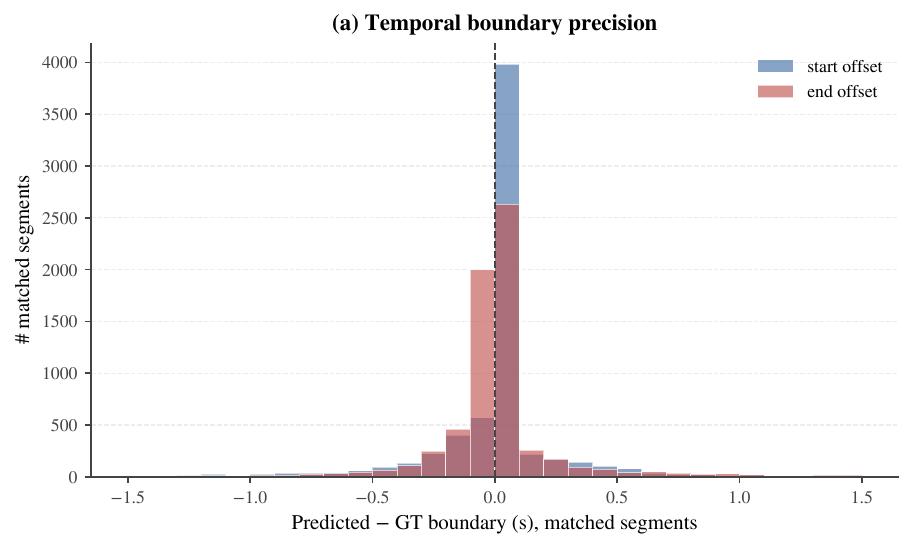}
\caption{\textbf{Boundary precision.} Signed offset between matched predicted and ground-truth segment boundaries.}
\label{fig:tprec}
\end{minipage}\hfill
\begin{minipage}[c]{0.44\textwidth}
\centering
\captionof{table}{\textbf{Segmentation and boundary statistics} of \method{}-8B on \dataset{}.}
\label{tab:tempstats}
\small\renewcommand{\arraystretch}{1.15}
\begin{tabular}{L{3.55cm} C{1.45cm}}
\toprule
Statistic & Value \\
\midrule
Mean GT segments / video & $\approx 2$ \\
Mean pred.\ segments / video & $\approx 1.9$ \\
Exact segment count & 65.7\% \\
\quad under- / over-seg. & 21.9\% / 12.4\% \\
Median $|\Delta|$ start / end & 0.00 / 0.04\,s \\
Within 0.5\,s (start / end) & 91.0 / 91.3\% \\
\bottomrule
\end{tabular}
\end{minipage}
\end{figure}

\paragraph{Temporal precision and segmentation behavior.}
For predicted segments matched at temporal IoU~$\geq0.5$, the boundaries are precise (Figure~\ref{fig:tprec}). The median absolute start and end offsets are $0.00$\,s and $0.04$\,s, and approximately $91\%$ of matched boundaries fall within $0.5$\,s of the annotation. At the clip level (Table~\ref{tab:tempstats}), \method{}-8B predicts approximately $1.9$ segments per video on average, compared with approximately $2$ in the ground truth. It predicts the exact number of segments for $65.7\%$ of clips, under-segments $21.9\%$, and over-segments $12.4\%$. The model is therefore slightly conservative: it is more likely to merge adjacent phases than to introduce spurious boundaries. This behavior is also visible in the qualitative examples in Section~\ref{app:analysis}.

\section{Qualitative Results}
\label{app:analysis}

\subsection{Representative Predictions}
\begin{figure}[p]
\centering
\includegraphics[width=.96\linewidth]{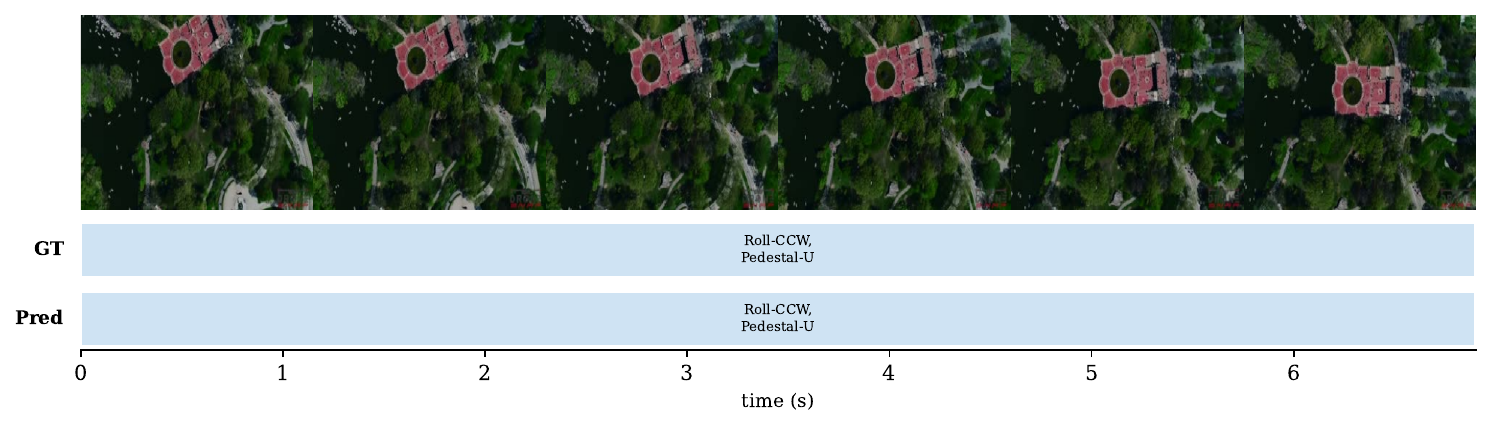}\\[4pt]\includegraphics[width=.96\linewidth]{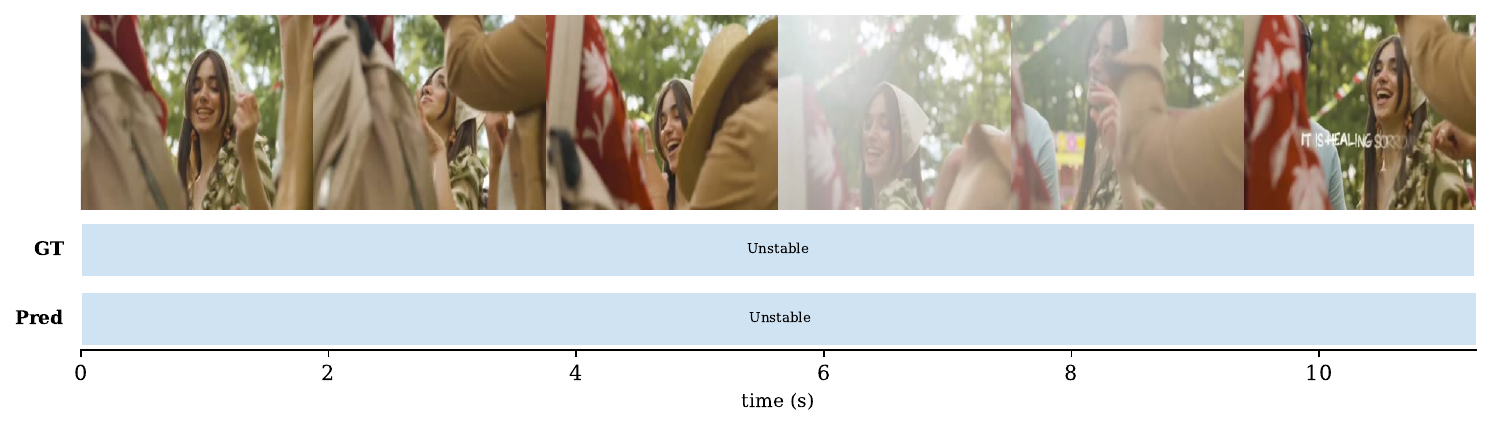}\\[4pt]\includegraphics[width=.96\linewidth]{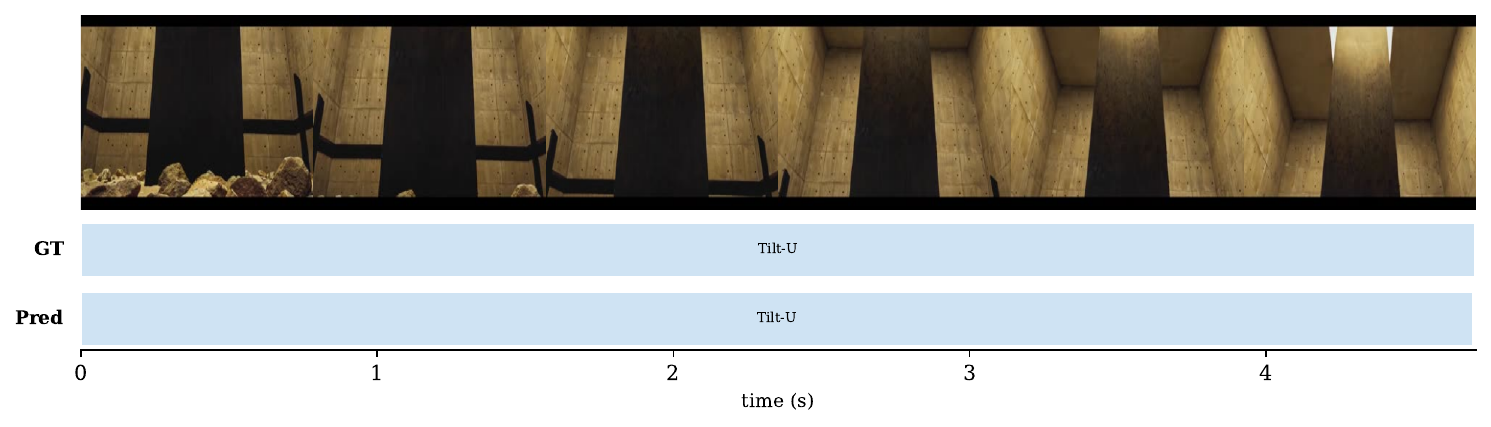}
\caption{\textbf{Qualitative predictions of \method{}-8B on \dataset{} (part~1).} Matched GT and Pred segments (by temporal IoU) share a color, unmatched segments are gray, and each segment lists its movements.}
\label{fig:qual_s1}
\end{figure}

\begin{figure}[p]
\centering
\includegraphics[width=.96\linewidth]{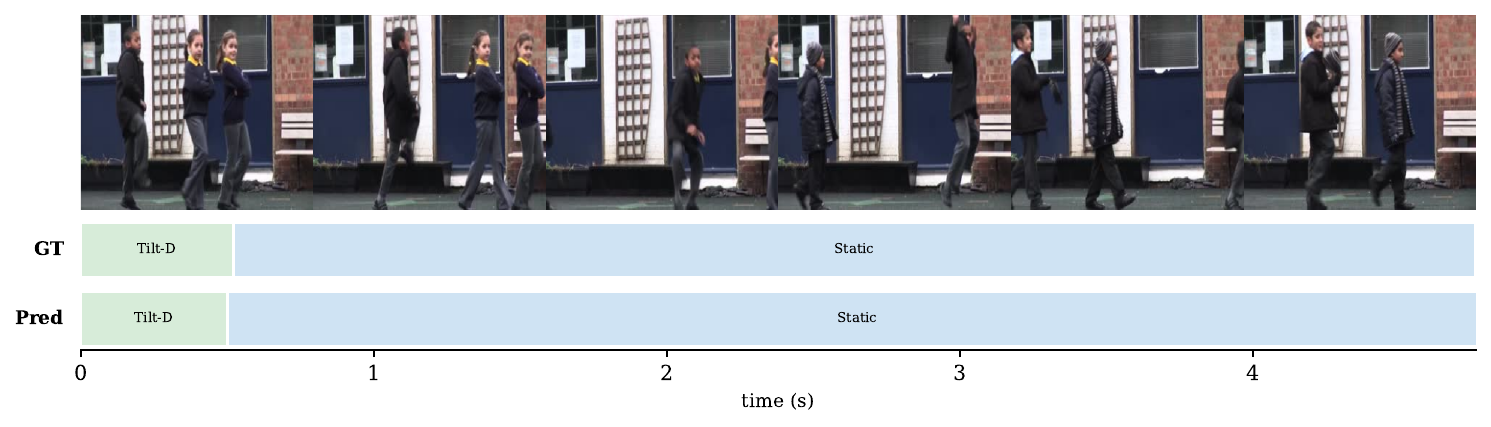}\\[4pt]\includegraphics[width=.96\linewidth]{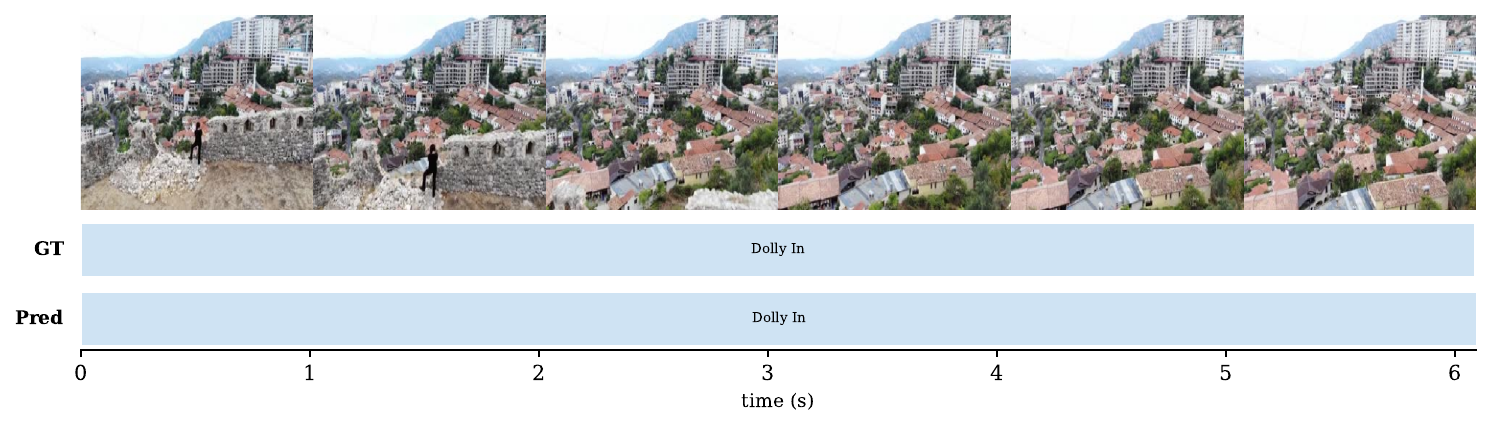}\\[4pt]\includegraphics[width=.96\linewidth]{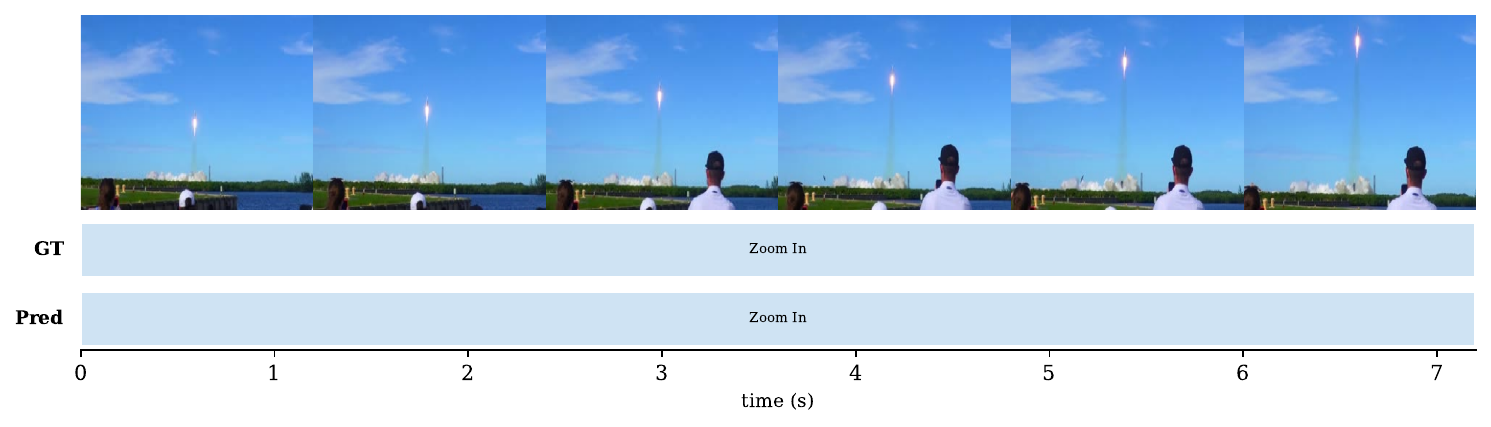}
\caption{\textbf{Qualitative predictions on \dataset{} (part~2).} Conventions as in Figure~\ref{fig:qual_s1}.}
\label{fig:qual_s2}
\end{figure}

\begin{figure}[p]
\centering
\includegraphics[width=.96\linewidth]{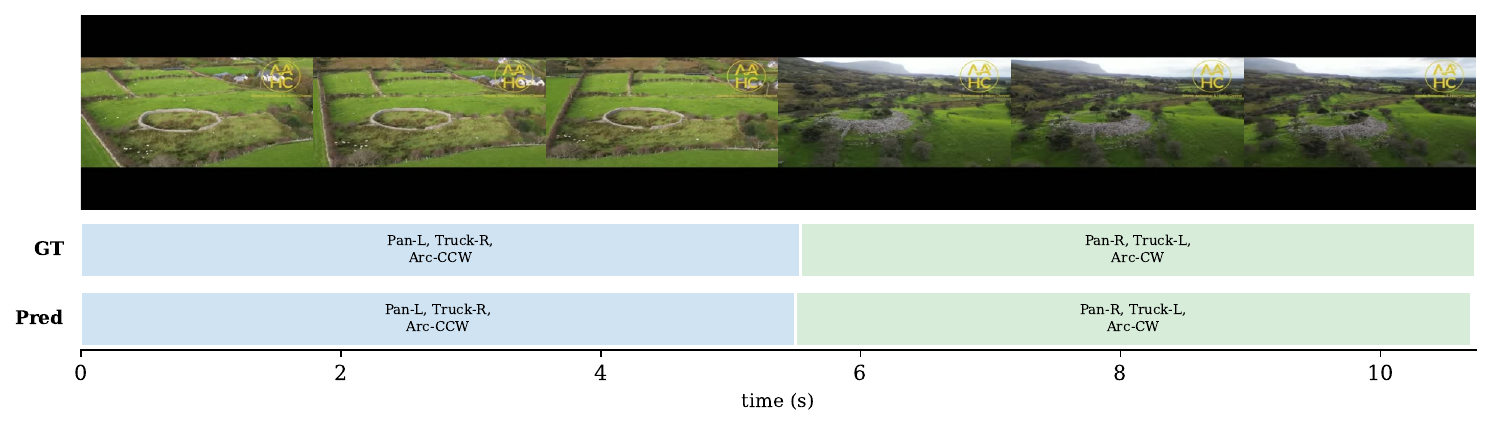}\\[4pt]\includegraphics[width=.96\linewidth]{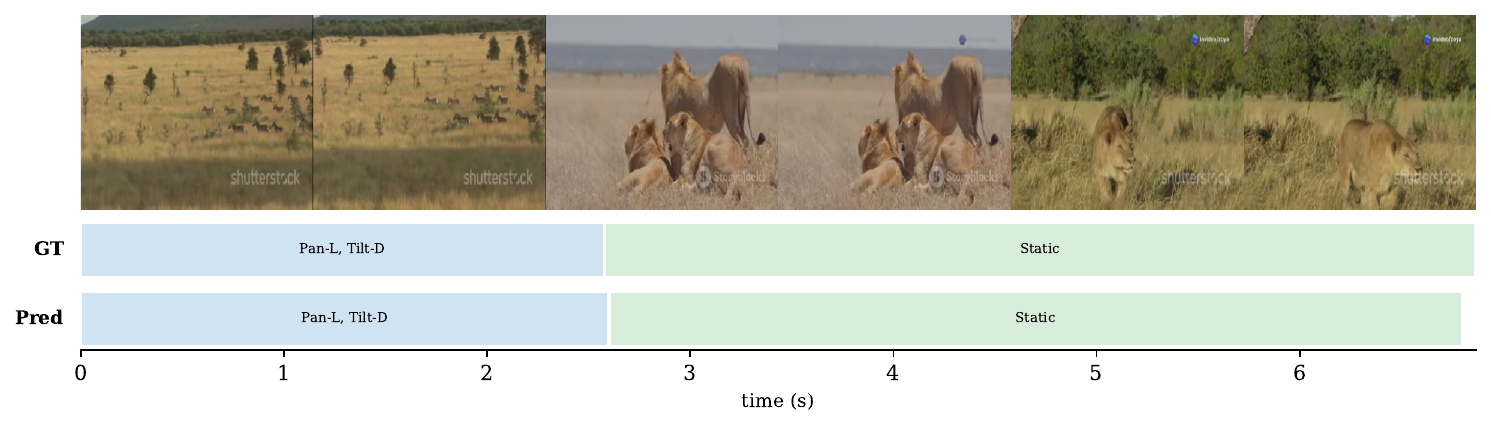}\\[4pt]\includegraphics[width=.96\linewidth]{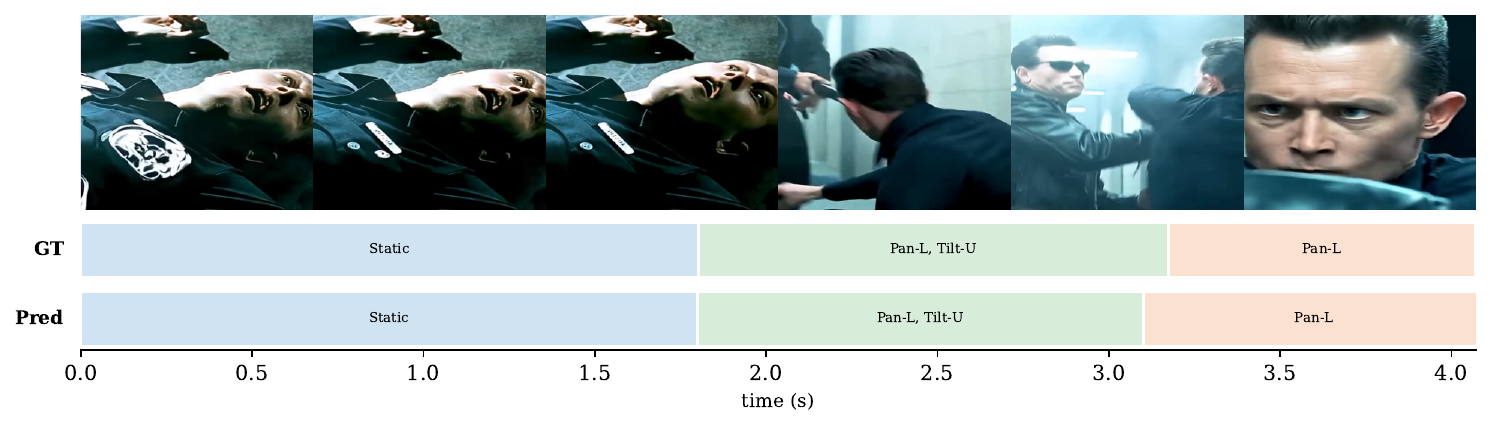}
\caption{\textbf{Qualitative predictions on \dataset{} (part~3).} Conventions as in Figure~\ref{fig:qual_s1}.}
\label{fig:qual_i1}
\end{figure}

\begin{figure}[p]
\centering
\includegraphics[width=.96\linewidth]{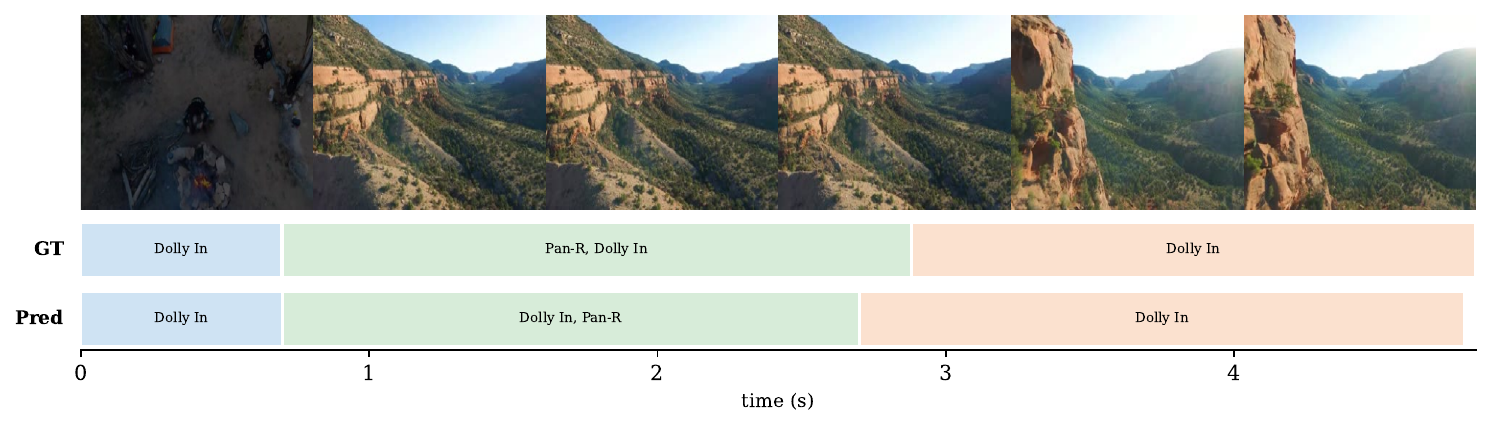}\\[4pt]\includegraphics[width=.96\linewidth]{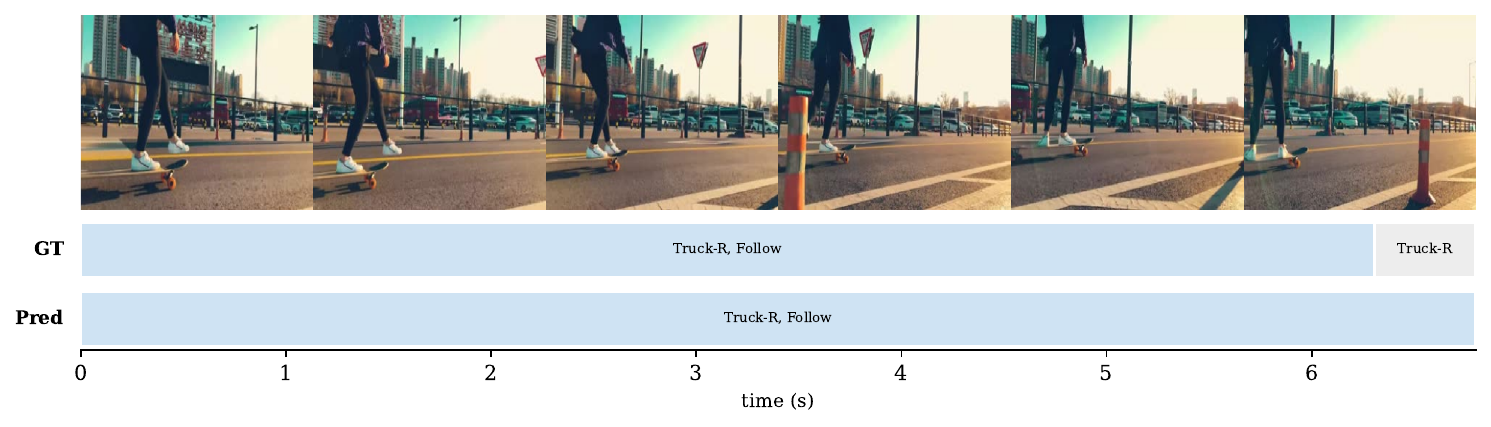}\\[4pt]\includegraphics[width=.96\linewidth]{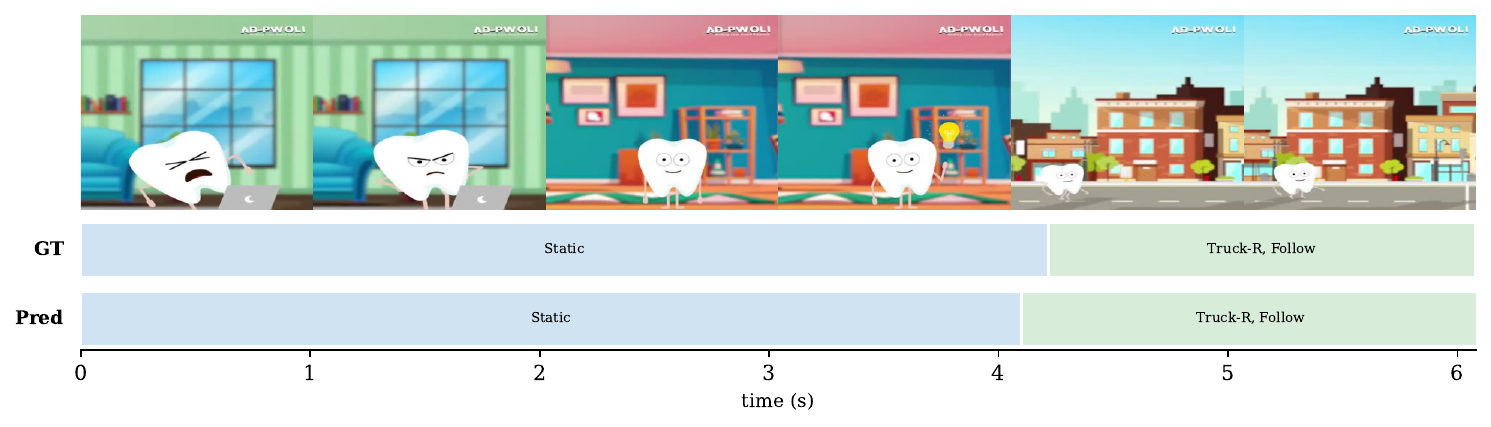}
\caption{\textbf{Qualitative predictions on \dataset{} (part~4).} Conventions as in Figure~\ref{fig:qual_s1}.}
\label{fig:qual_i2}
\end{figure}

\begin{figure}[p]
\centering
\includegraphics[width=.96\linewidth]{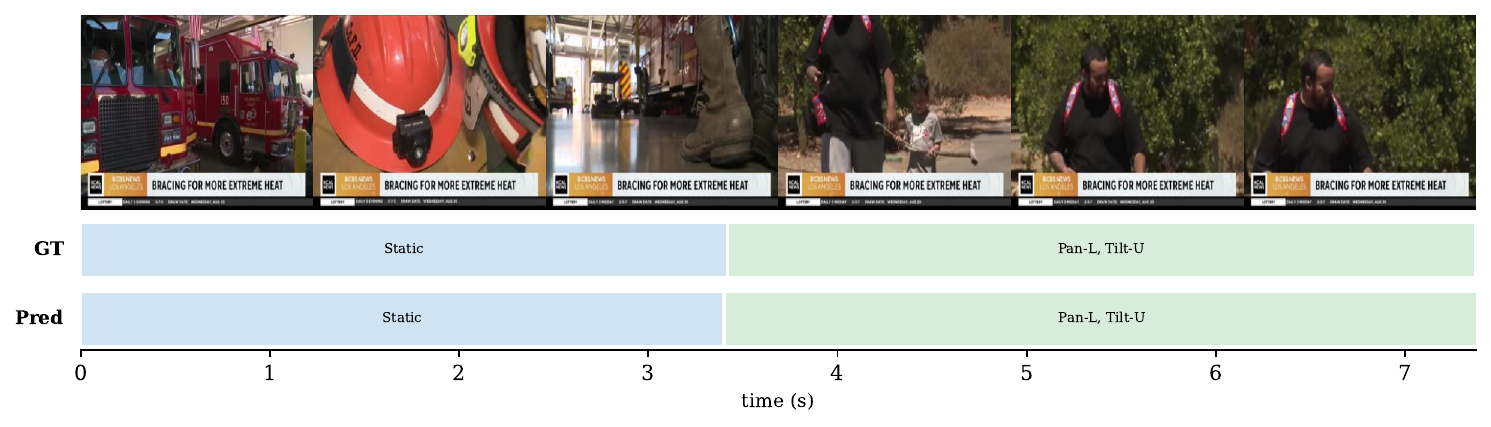}\\[4pt]\includegraphics[width=.96\linewidth]{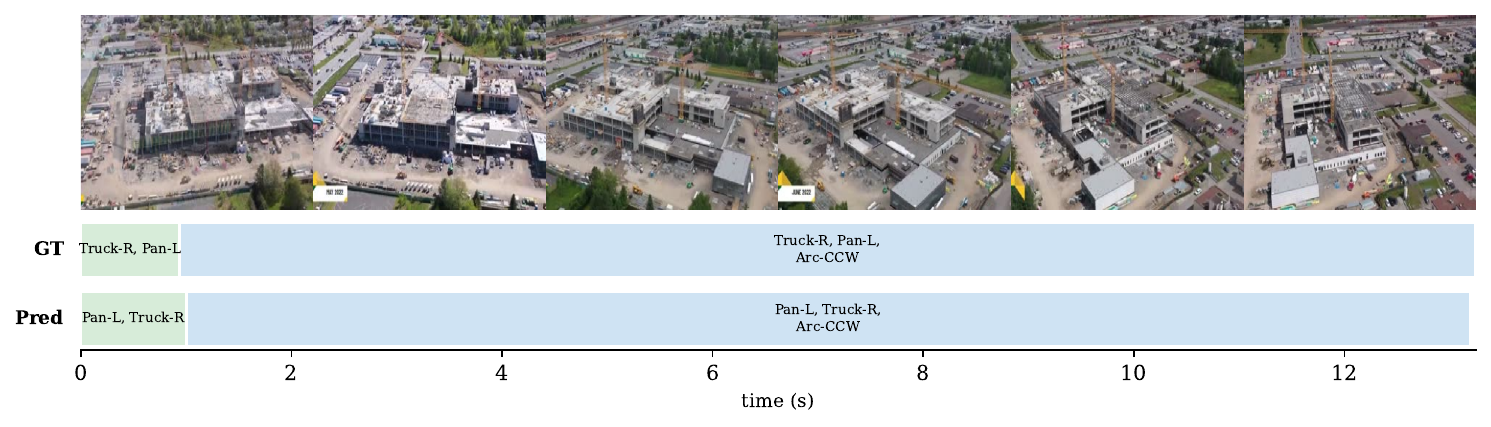}\\[4pt]\includegraphics[width=.96\linewidth]{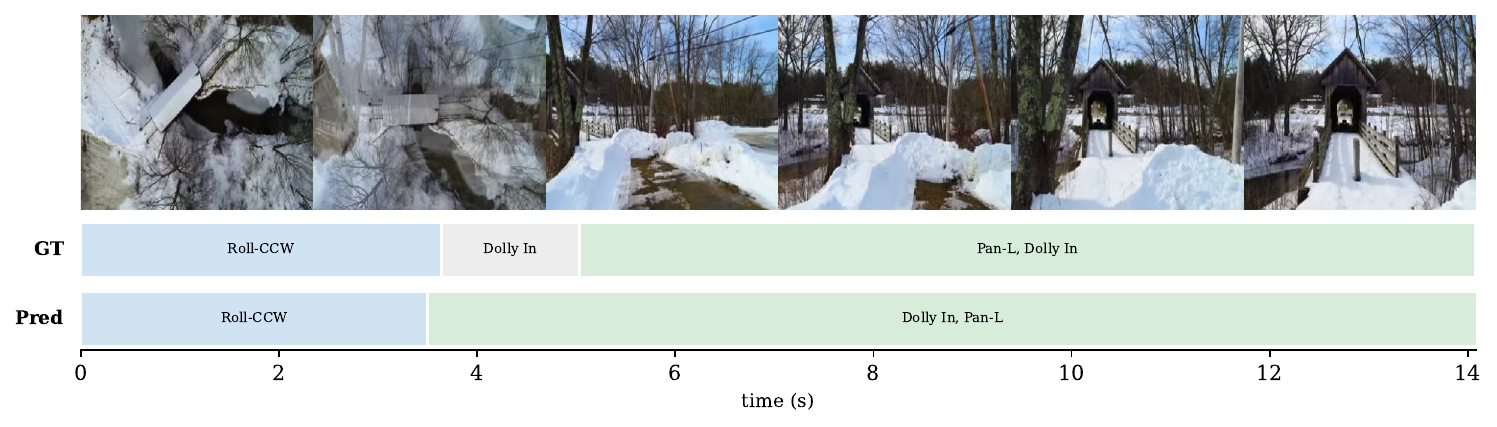}\\[4pt]\includegraphics[width=.96\linewidth]{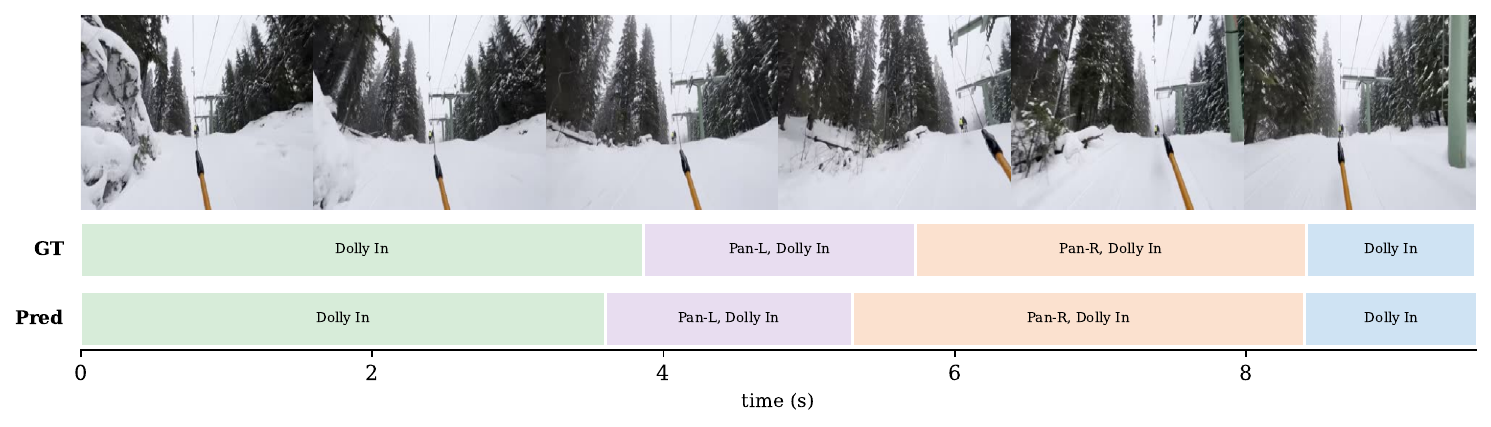}
\caption{\textbf{Qualitative predictions on \dataset{} (part~5).} Conventions as in Figure~\ref{fig:qual_s1}.}
\label{fig:qual_c1}
\end{figure}

\begin{figure}[p]
\centering
\includegraphics[width=.96\linewidth]{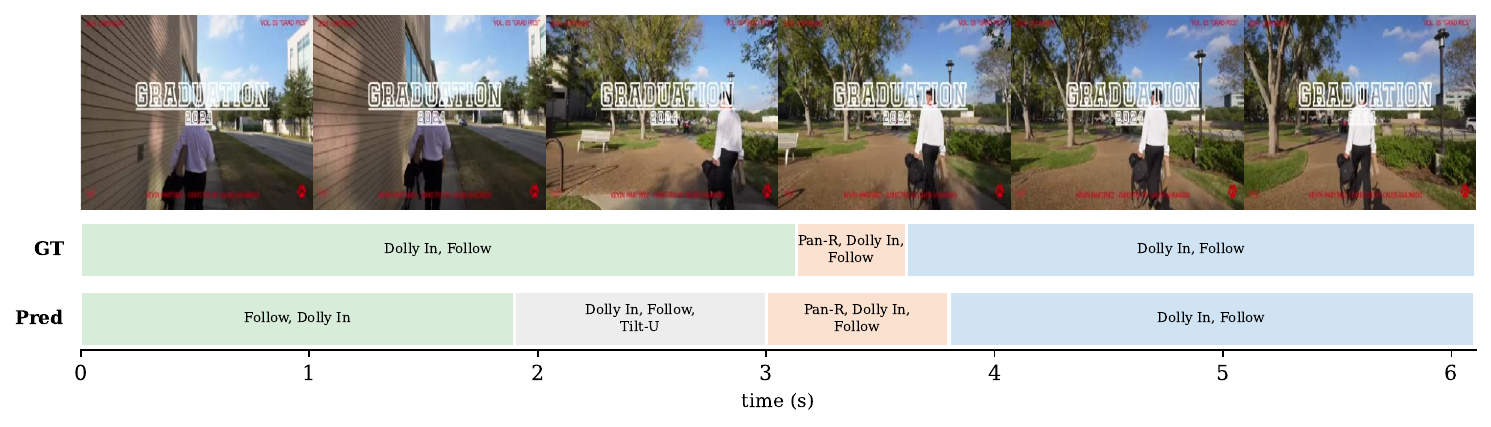}\\[4pt]\includegraphics[width=.96\linewidth]{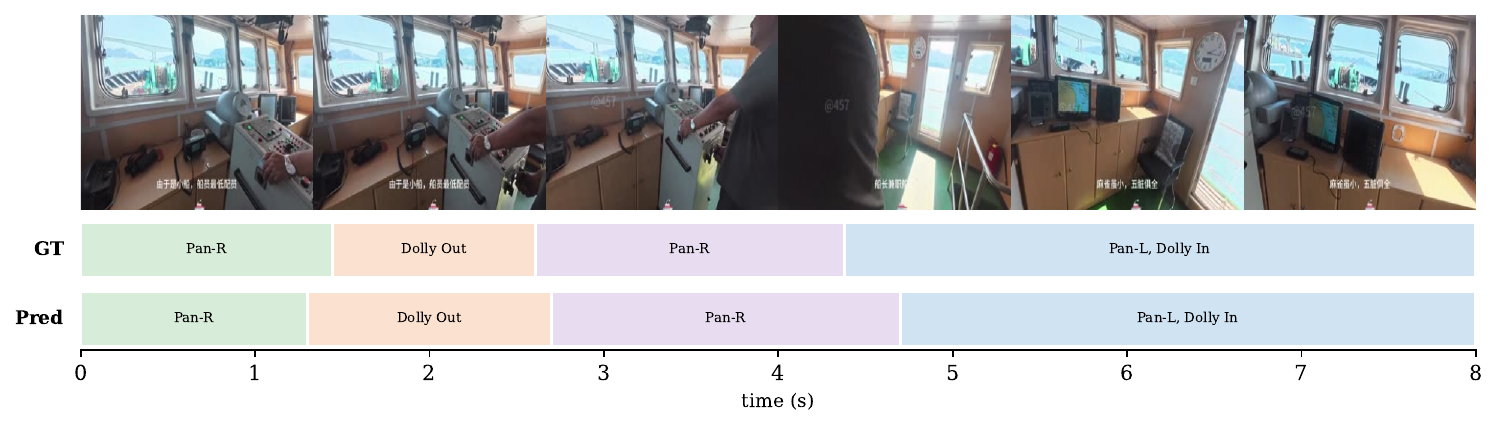}\\[4pt]\includegraphics[width=.96\linewidth]{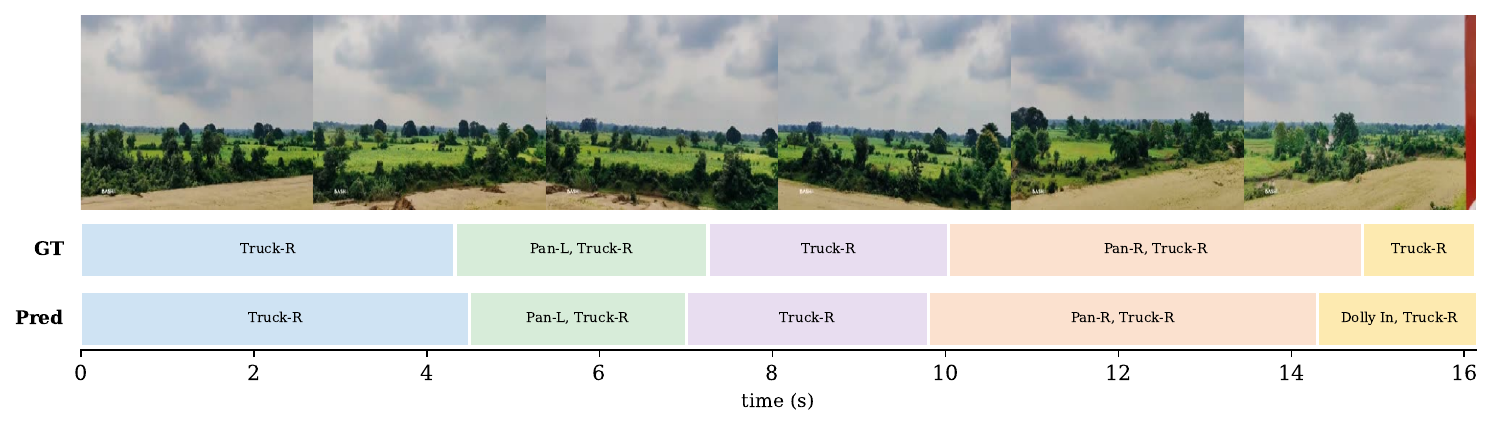}\\[4pt]\includegraphics[width=.96\linewidth]{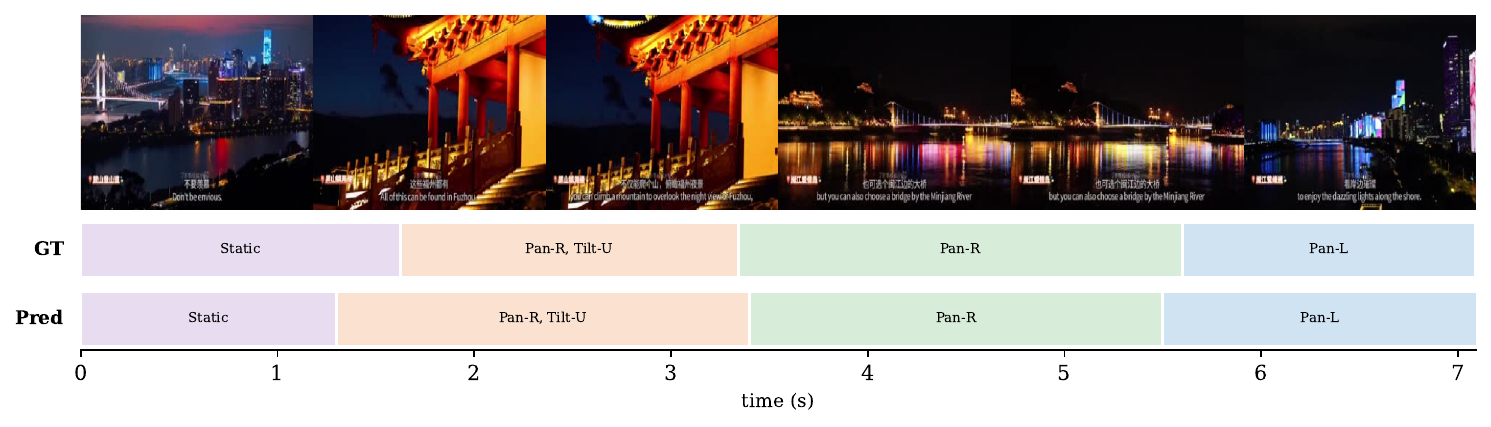}
\caption{\textbf{Qualitative predictions on \dataset{} (part~6).} Conventions as in Figure~\ref{fig:qual_s1}.}
\label{fig:qual_c2}
\end{figure}

Figures~\ref{fig:qual_s1}--\ref{fig:qual_c2} show example predictions of \method{}-8B on \dataset{}, drawn from clips that span a range of temporal and compositional complexity. In each panel, sampled frames appear above the ground-truth (GT) and predicted (Pred) segment timelines. Segments matched by temporal IoU share a color, unmatched segments are gray, and every segment is labeled with its movements. Across these examples, the model recovers the dominant movements and the overall temporal structure, including compositional cases such as a simultaneous Roll and Pedestal Up or a Truck$+$Pan$+$Arc orbit. Its remaining errors are mostly merged adjacent phases or a movement dropped from a densely compositional segment, consistent with the error analysis in Appendix~\ref{app:error_analysis}.

\subsection{Failure Cases}
\begin{figure}[t]
\centering
\includegraphics[width=.96\linewidth]{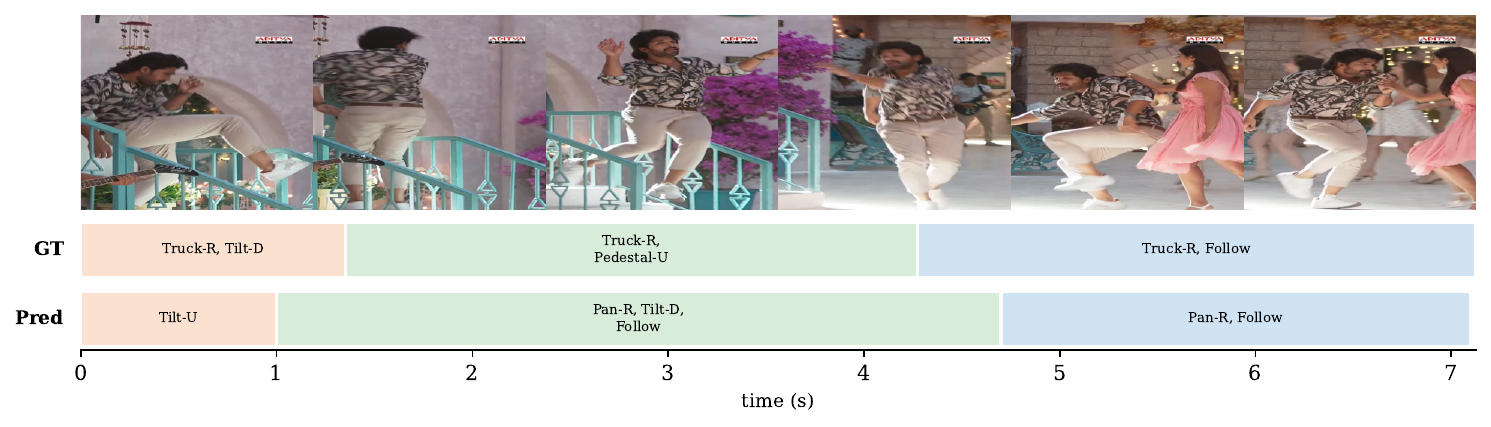}\\[4pt]\includegraphics[width=.96\linewidth]{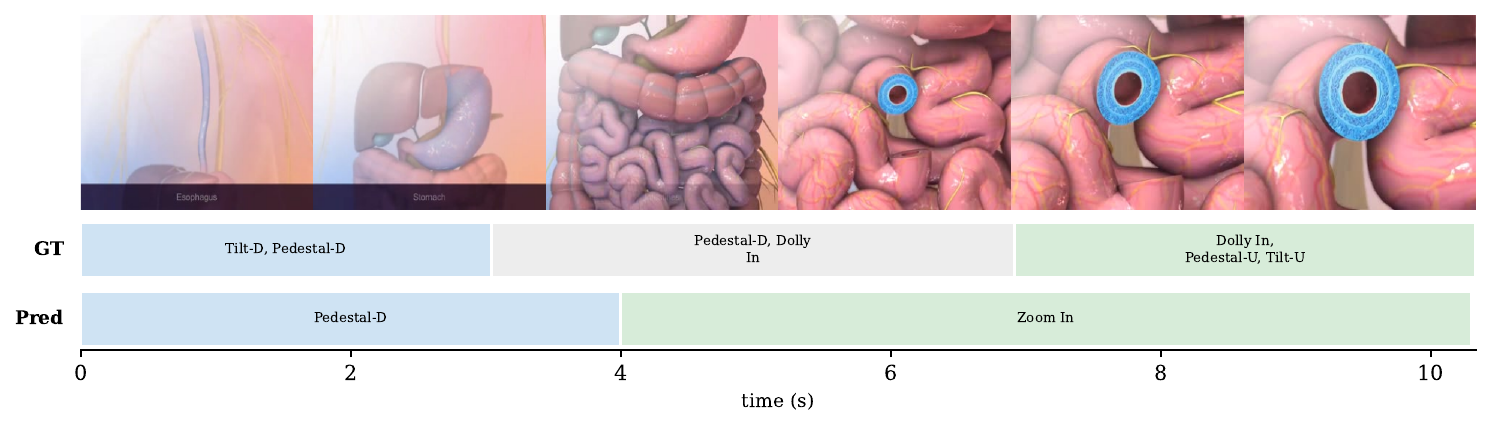}\\[4pt]\includegraphics[width=.96\linewidth]{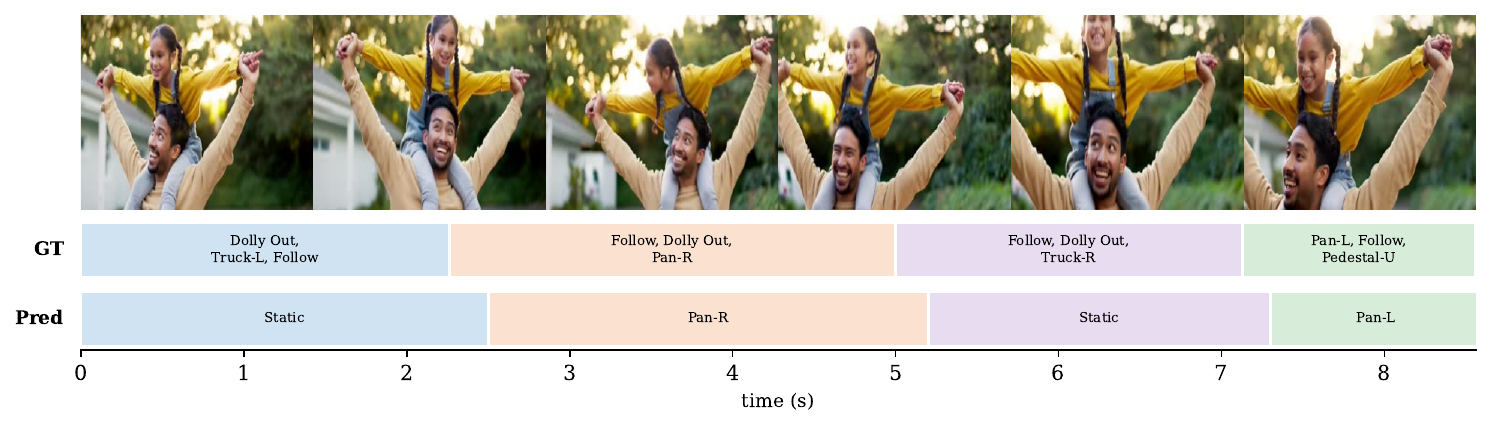}\\[4pt]\includegraphics[width=.96\linewidth]{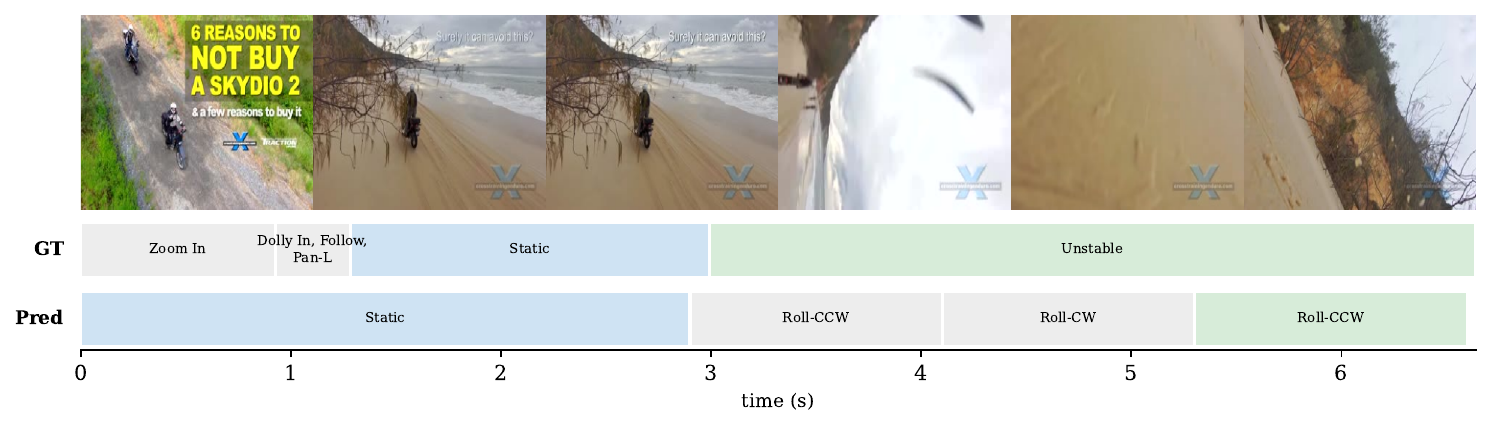}
\caption{\textbf{Failure cases of \method{}-8B.} Top to bottom: a lateral Truck predicted as Pan; a forward Dolly predicted as Zoom; a hand-held Follow predicted as Static or Pan; and Unstable footage predicted as Roll and over-segmented into short intervals. Conventions as in Figure~\ref{fig:qual_s1}.}
\label{fig:qual_fail}
\end{figure}

\method{} still breaks down on the hardest clips, and Figure~\ref{fig:qual_fail} shows four representative errors. They fall into a few recurring modes. First, the model confuses movements that look similar in the image plane but differ geometrically, reading a lateral Truck as a Pan and a forward Dolly as a Zoom; both distinctions require the parallax cues that its encoder captures poorly. Second, it misses subject-referenced motion, labeling a hand-held Follow shot as Static or a plain Pan. Third, it mistakes irregular Unstable footage for Roll and over-segments it into several short intervals. Across all four, dense compositions are only partially recovered and adjacent phases are often merged. These modes match the quantitative error analysis in Appendix~\ref{app:error_analysis}, where most residual errors are incomplete or incorrect label sets rather than boundary errors.

\end{document}